\documentclass[10pt,paper,twoside,web]{IEEEtran}

\usepackage{generic}
\usepackage{color}
\usepackage{cite}
\usepackage{amsmath,amssymb,amsfonts}
\usepackage{algorithmic}
\usepackage{graphicx}
\usepackage{textcomp}
\usepackage{subfigure}
\usepackage{ulem}
\usepackage{caption}
\usepackage{float} 
\usepackage{subcaption}
\usepackage[linesnumbered,ruled,vlined]{algorithm2e} %

\newcommand{\eproof}{\hfill\rule{2mm}{2mm}}

\newcommand{\bstate}{\medskip\begin{state} }
	\newcommand{\estate}{ \hfill  \rule{1mm}{2mm}\medskip\end{state}}

\newcommand{\bass}{\medskip\begin{ass} }
	\newcommand{\eass}{ \hfill  \rule{1mm}{2mm}\medskip\end{ass}}

\newcommand{\brem}{\medskip \begin{remark}  }
	\newcommand{\erem}{\hfill \rule{1mm}{2mm}\medskip
\end{remark} }
\newcommand{\bthm}{\medskip\begin{theorem}  }
	\newcommand{\ethm}{ \hfill  \rule{1mm}{2mm} \medskip
\end{theorem} }
\newcommand{\blem}{\medskip\begin{lemma}  }
	\newcommand{\elem}{ \hfill \rule{1mm}{2mm}\medskip
\end{lemma} }
\newcommand{\bcorollary}{\medskip\begin{corollary}  }
	\newcommand{\ecorollary}{  \hfill \rule{1mm}{2mm}\medskip
\end{corollary} }
\newcommand{\bdefn}{\medskip\begin{definition}}
	\newcommand{\edefn}{  \hfill \rule{1mm}{2mm}\medskip
\end{definition} }
\newcommand{\bproposition}{\medskip\begin{proposition} }
	\newcommand{\eproposition}{\hfill \rule{1mm}{2mm}\medskip
\end{proposition} }
\newcommand{\bexample}{\medskip\begin{example} \rm}
	\newcommand{\eexample}{ \hfill \rule{1mm}{2mm}\medskip
\end{example} }

\newcommand{\bcon}{\medskip\begin{condition} \rm}
	\newcommand{\econ}{ \hfill \rule{1mm}{2mm}\medskip
\end{condition} }

\renewcommand{\t}{^\top}
\newcommand{\proofnow}{\noindent{\bf Proof: }}

\newtheorem{theorem}{\bf Theorem}[section]
\newtheorem{ass}{\bf Assumption}[section]
\newtheorem{lemma}{\bf Lemma}[section]
\newtheorem{definition}{\bf Definition}[section]
\newtheorem{remark}{\bf Remark}[section]
\newtheorem{corollary}{\bf Corollary}[section]
\newtheorem{proposition}{\bf Proposition}[section]
\newtheorem{example}{\bf Example}[section]
\newtheorem{condition}{\bf Condition}[section]
\newtheorem{state}{\bf Assumption}[section]

\begin{document}
\title{A Gait Driven Reinforcement Learning Framework for Humanoid Robots } 
\author{Bolin Li, Yuzhi Jiang, Linwei Sun, Xuecong Huang, Lijun Zhu, and Han Ding}


\maketitle
\begin{abstract}
This paper presents a real-time gait driven training framework for humanoid robots. First, we introduce a novel gait planner that incorporates dynamics to design the desired joint trajectory. In the gait design process, the 3D robot model is decoupled into two 2D models, which are then approximated as hybrid inverted pendulums (H-LIP) for trajectory planning. The gait planner operates in parallel in real time within the robot's learning environment. Second, based on this gait planner, we design three effective reward functions within a reinforcement learning framework, forming a reward composition to achieve periodic bipedal gait. This reward composition reduces the robot’s learning time and enhances locomotion performance. Finally, a gait design example, along with simulation and experimental comparisons, is presented to demonstrate the effectiveness of the proposed method.
\end{abstract}


\section{Introduction}
The development of stable and efficient bipedal locomotion remains a fundamental challenge in humanoid robotics, particularly in dynamic and unstructured environments\cite{thor2020generic}. In general, a specific gait can be viewed as a dynamic process that has a characteristic periodic structure, but is also able to flexibly adapt to moderate environment disturbances \cite{he2025attention}.

Traditional methods for gait generation rely heavily on model-based optimization \cite{wieber2006online} or heuristic rules \cite{raibert1986legged}, such as joint trajectory optimization for stability enhancement \cite{wieber2006online} or gravity-aligned foothold heuristics for legged locomotion \cite{raibert1986legged, grandia2023perceptive}. Although these methods provide analytical guarantees, they often lack adaptability to precious dynamics or suffer from high computational complexity, limiting their practicality in unstructured settings.

Recent advances in reinforcement learning (RL) have introduced promising alternatives by enabling robots to learn locomotion policies through environmental interaction \cite{he2025attention,lee2020learning,jenelten2024dtc}. In \cite{peng2021amp}, reward composition frameworks and adversarial imitation learning techniques have demonstrated the ability to generate lifelike motions without manual objective design. In \cite{siekmann2021sim}, a reward-specification framework was proposed for sim-to-real reinforcement learning of bipedal locomotion, using probabilistic periodic costs to define intuitive, parametric reward functions for common gaits, enabling successful transfer of learned gaits to the Cassie robot and a generic policy for gait transitions. However, purely learning-based approaches face significant limitations, including poor interpretability \cite{sheng2025comprehensive} and prolonged training times. Moreover, RL-generated motions frequently exhibit irregular periodicity, hindering their deployment on physical platforms where stability and rhythmicity are critical.

A key research gap exists in bridging the divide between model-based robustness and data-driven adaptability. Probabilistic model-based reinforcement learning (MBRL) offers a promising solution to enhance sample efficiency and generalizability. It involves learning a data-driven dynamics model of a robot and performing various skills using probabilistic model-predictive control with varying control objectives. However, due to the computational expense of MBRL \cite{moerland2023model}, previous studies on biped locomotion were primarily conducted with physical robots using offline planning \cite{deisenroth2012toward}. Although \cite{kuo2023reinforcement} proposed a probabilistic model-based reinforcement learning approach to learn the energy exchange dynamics of a spring-loaded biped robot, demonstrating successful on-site walking acquisition with a reduced-order dynamics model, energy exchange dynamics, and 40 Hz real-time planning, it only adopted a reduced-order dynamics model without considering the planning of the whole-body joint trajectory.

While hybrid approaches integrating intrinsic models with RL have emerged, existing solutions often fail to balance real-time planning whole-body joint trajectory efficiency with the learning process’s exploratory requirements. Model-based controllers, despite their dynamical fidelity, struggle with real-time trajectory synthesis for high-dimensional humanoid systems \cite{xu2023robust}, whereas RL policies require excessive computational resources to converge to periodic gaits.

To tackle these challenges, this paper introduces a novel framework that integrates real-time gait planning with structured reward composition in RL. Our approach features two key innovations: (1) a dynamic-aware gait planner that decomposes the 3D humanoid model into two decoupled 2D subsystems, approximated as H-LIP proposed in \cite{9723475}, enabling real-time high-dimensional joint trajectory generation while preserving crucial dynamical properties; and (2) a structured reward composition strategy that systematically combines periodicity enforcement, trajectory tracking, and time efficiency metrics to guide policy learning. By merging analytical model simplifications with data-driven adaptation, our method ensures efficient training convergence and robust locomotion performance.

The contributions of this work are threefold:

(1) A dynamical decoupling strategy that simplifies 3D humanoid motion into computationally tractable 2D H-LIP approximations, generating feasible reference trajectories under balance and kinematic constraints.

(2) A multi-objective reward design that explicitly promotes periodic, time-efficient, and tracking-consistent gaits, addressing common RL challenges such as local optima and erratic behaviors.

(3) Extensive benchmarking against baseline methods, showcasing superior learning efficiency and locomotion quality in simulated environments, with promising implications for deployment on physical humanoid platforms.

This work advances the integration of model-based planning and learning-based control, offering a scalable solution for bipedal locomotion in learning environments. By addressing the dual challenges of real-time adaptability and motion periodicity, our framework paves the way for more reliable and training-efficient humanoid robots capable of operating in real-world dynamic settings. This letter is organized as follows. Section \ref{sec:problem_formulation} outlines the preliminaries; Section \ref{sec:Modelling} presents the modeling of the humanoid robot; Section \ref{sec:gait_design} introduces the gait design; Section \ref{se:reward_composition} discusses the learning framework; Section \ref{sec:design_example} provides a design example and evaluates the performance of the proposed framework. Finally, Section \ref{sec:conclusion} concludes the paper.

\section{Preliminaries}\label{sec:problem_formulation}

\begin{figure}[!ht]
	\centering
	\includegraphics[scale=0.35]{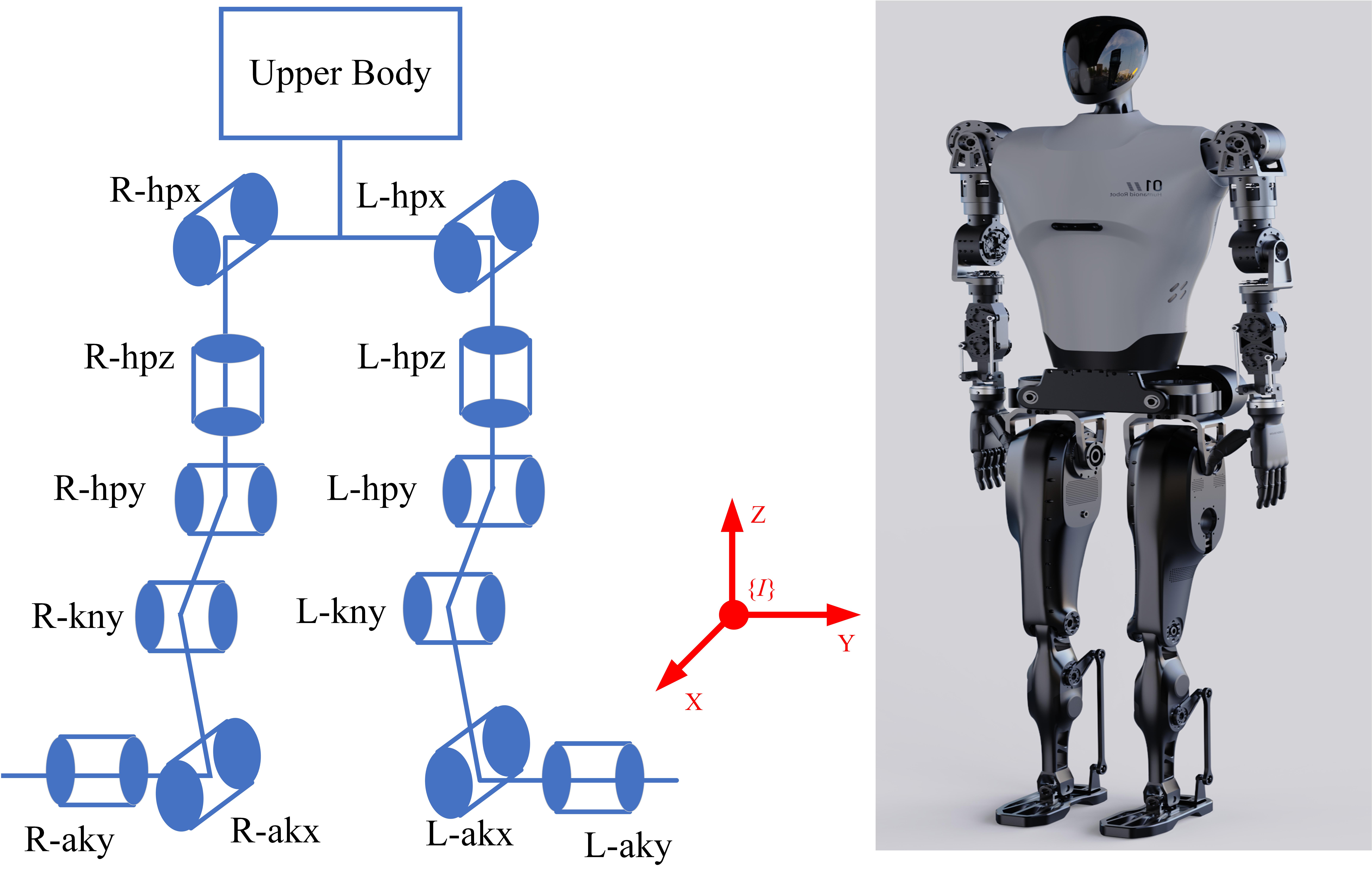}
	\caption{Motor distribution (left) and mechanical structure of the humanoid robot (right).}
	\label{fig_mechanical_structure}
\end{figure}

\begin{figure*}[htbp]
	\centering
	\includegraphics[scale=0.40]{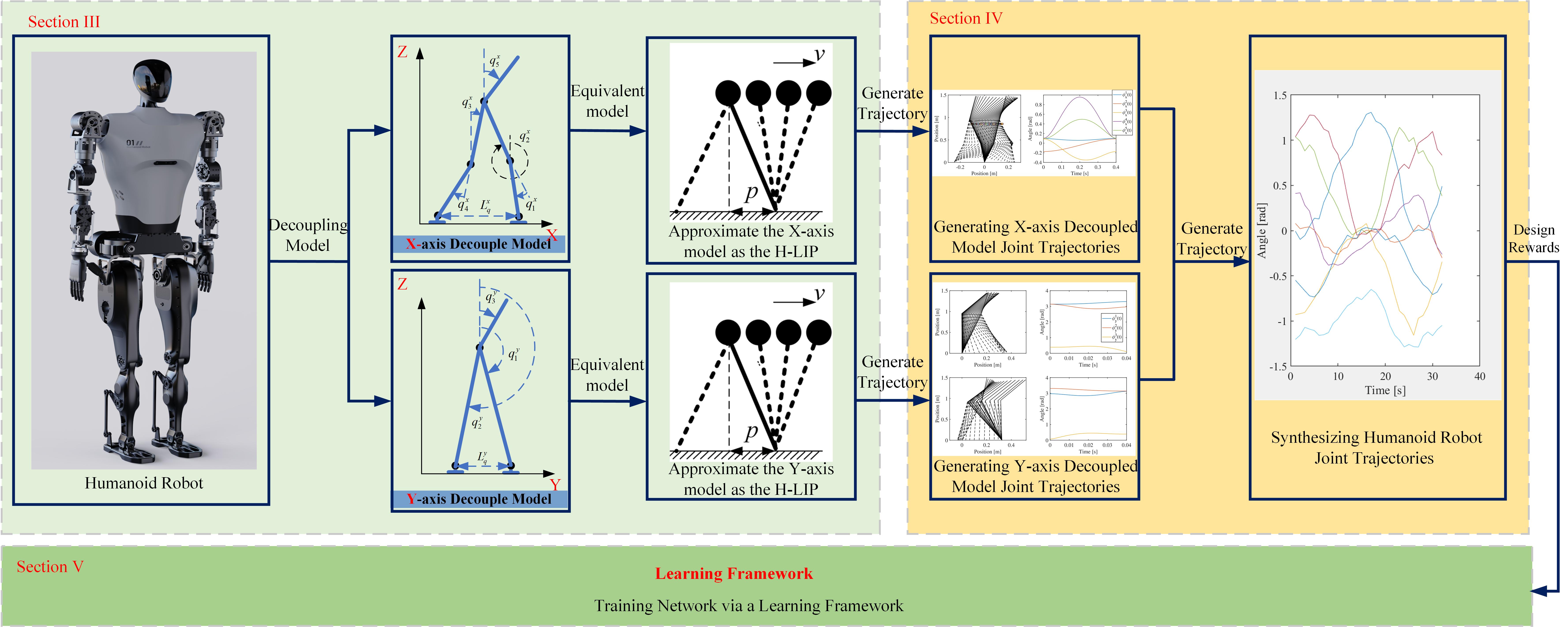}
	\caption{A real-time-gait-driven training framework.}
	\label{fig_pipline}
\end{figure*}

The humanoid robot studied in this paper, along with its mechanical structure and motor distribution, is shown in Fig. \ref{fig_mechanical_structure}. It features twelve joints in the lower body of the mechanical structure. The ankle joints are parallel joints, and the foot pitch angle ($q_{\rm{L-aky}}$ and $q_{\rm{R-aky}}$  as well as the foot roll angle $q_{\rm{L-akx}}$ and $q_{\rm{R-akx}}$) can be derived from the angles of the parallel joints \cite{liang2025reduced}. The locomotion of typical humanoid robots with the same mechanical configuration has been studied in \cite{xiong2020dynamic, orin2013centroidal}. In this paper, we propose a reinforcement learning framework integrated with a real-time gait planner, which uses robot dynamics as part of the reward function to accelerate learning and achieve periodic bipedal locomotion. The proposed training framework driven by real-time gait generation and consisting of gait design and network training is illustrated in Fig. \ref{fig_pipline}. First, we decouple the robot model into separate $X$-axis and $Y$-axis models, which are approximated as a H-LIP to generate joint trajectories for each decoupled model. These trajectories are then combined to form the final joint trajectories for the full robot model. Finally, within the robot locomotion learning framework, we design the reward function by incorporating the joint trajectories of the robot model.

\section{Modelling of the Humanoid Robot}\label{sec:Modelling}

In this section, we begin by decoupling the robot model into two planar models and then introduce the H-LIP model. The H-LIP model will serve as an approximation of the two planar model in the subsequent gait planner design.

\begin{figure}[!ht]
	\centering
	\includegraphics[scale=0.72]{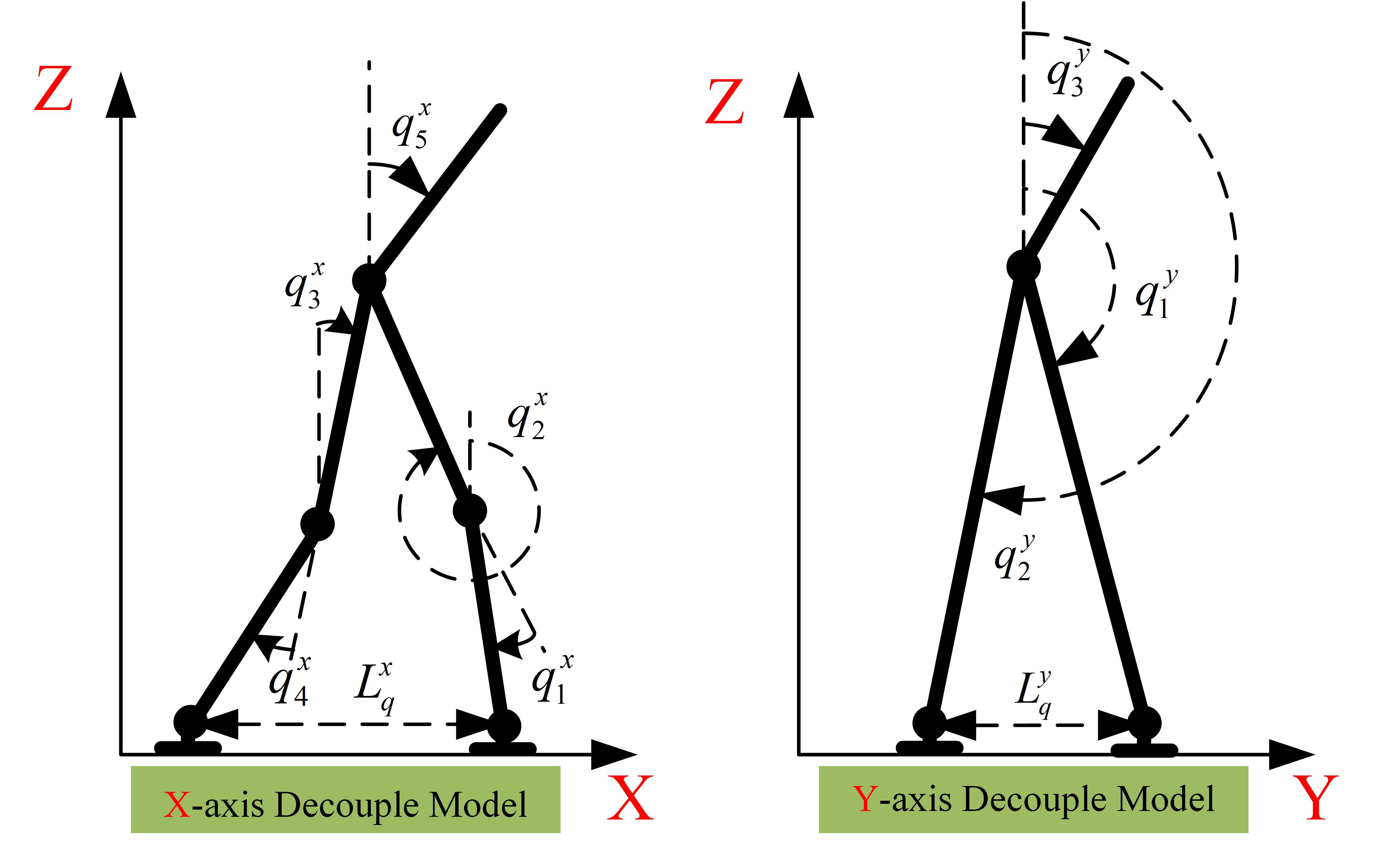}
	\caption{Decouple model from the mechanical structure.}
	\label{fig_decouple_model}
\end{figure}

\subsection{Robot Model Decoupling}
Assume that the mass of the hands is minimal, and the upper body is treated as a rigid body in this paper. Similar to \cite{zhao2023compliant}, we decouple the humanoid robot into a five-link planar robot along the $x$-axis ($X$-model) and a three-link planar robot along the $y$-axis ($Y$-model). The decoupled model is illustrated in Fig. \ref{fig_decouple_model}. As shown in Fig. \ref{fig_decouple_model}, we define the generalized coordinates $q^x = [q^x_i] \in \mathbb{R}^5$ in $X$-model and $q^y=[q^y_i] \in \mathbb{R}^3$ in $Y$-model to design the corresponding joint trajectory. These absolute generalized coordinates are provided in both the $X$-model and $Y$-model for the simplified design of the subsequent gait planner. In fact, the dynamics of the $X$-model are influenced by $q^y$, and the dynamics of the $Y$-model are influenced by $q^x$. However, we assume that the influence in the orthogonal direction is minimal, as noted in \cite{xiong20223}. Given the $q_{\rm{R-hpz}}$ and $q_{\rm{L-hpz}}$ small angle changes (see Fig. \ref{fig_mechanical_structure}) of the humanoid robot, their effect on the position and velocity of the center of mass (CoM) in both the $X$-model and $Y$-model is negligible during the robot's walking motion. Thus, we assume the angles $q_{\rm{R-hpz}}$ and $q_{\rm{L-hpz}}$ to be zero.

\subsection{H-LIP model}


\begin{figure}[!ht]
	\centering
	\includegraphics[scale=0.55]{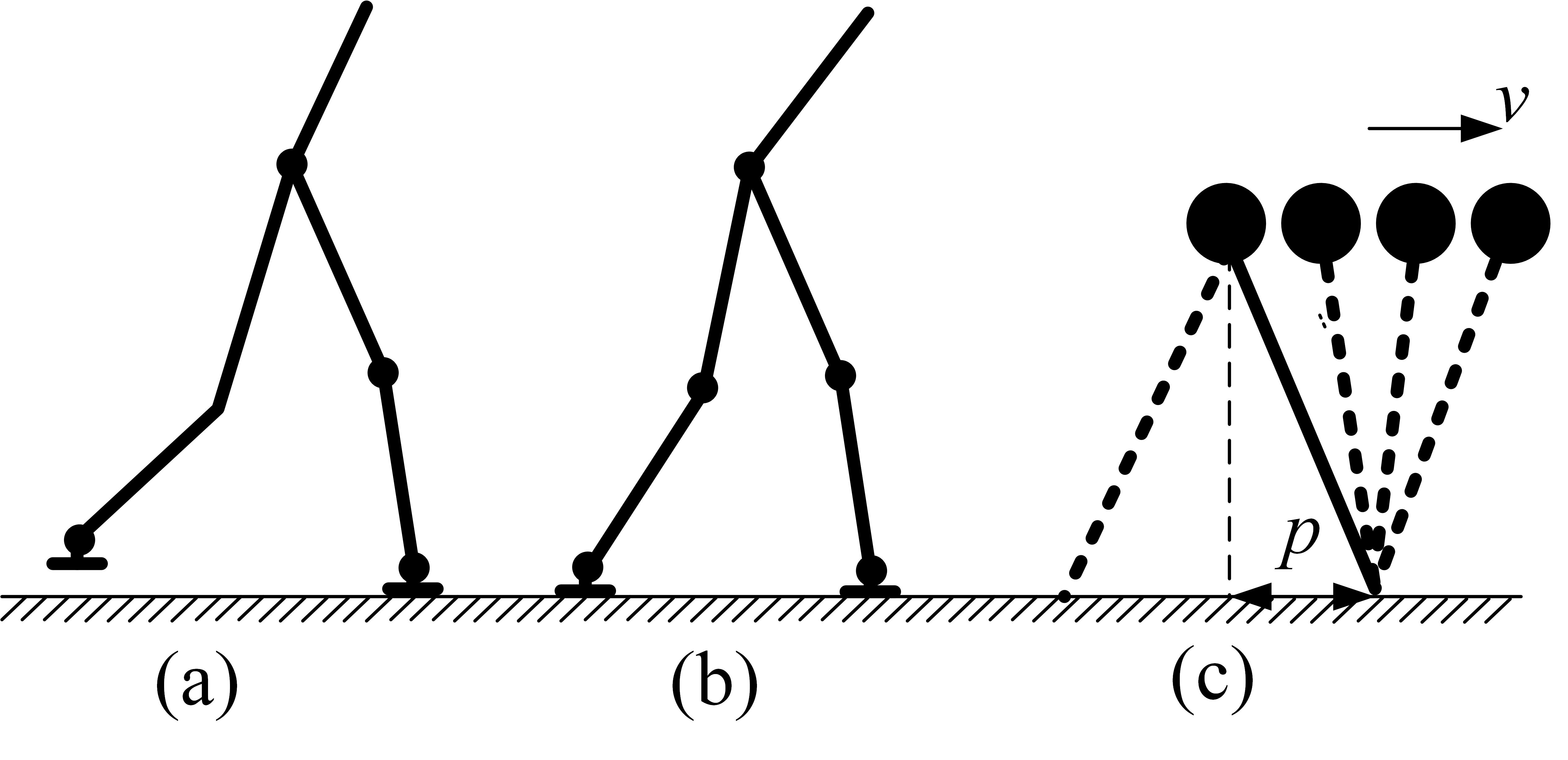}
	\caption{(a) Single suppose phase; (b) Double support phase; (c) H-LIP.}
	\label{fig_SSP_DSP}
\end{figure}
When the robot walks in the ground, the legs will contact with the ground. Based on the number of the contact point with ground, the walking is composed of a single support phase (SSP) as is shown in (a) of Fig. \ref{fig_SSP_DSP} and a double support phase (DSP) as is shown in (b) of Fig. \ref{fig_SSP_DSP}. The H-LIP is a point-mass model with a constant CoM height and two telescopic legs with point-feet as is shown in (c) of Fig. \ref{fig_SSP_DSP}.  In the SSP, the model is a passive LIP with no actuation. In this article, we assume that the DSP is instantaneous and occurs when the swing leg makes contact with the ground. The time interval between the previous DSP and the current DSP is referred to as the swing period. Due to the exchange between SSP and DSP, the H-LIP dynamics form a hybrid system, consisting of continuous dynamics and state transitions. Denote position $p \in \mathbb{R}$ representing the distance between the CoM and the support leg and velocity $v \in \mathbb{R}$ representing the velocity of the CoM. The continuous dynamics in H-LIP dynamics, called SSP dynamics, are \cite{xiong20223} 
\begin{align}
	\quad \ddot p =& {\lambda ^2}p \label{eq:SSP_dynamics}
\end{align}
where $\lambda  = \sqrt {\frac{g}{{{z_0}}}}$, $g$ is  gravitational acceleration, and $z_0$ is the height of the H-LIP. Let  $x = [p, v]\t$. The state-space representation of the continuous dynamics (\ref{eq:SSP_dynamics}) is
\begin{gather}
	{\dot {x}} = {A_{\rm{SSP}}}{x} \label{eq:state_space_representation}
\end{gather}  
where 
\begin{gather}
	{A_{\rm{SSP}}} = \left[ {\begin{array}{*{20}{c}}
			0&{1}\\
			\lambda^2&0
	\end{array}} \right].
\end{gather}

When the robot contact the ground the position $p$ and velocity $v$ is transited. 
Let $p^+$ and $p^-$ denote the position of the CoM after the swing leg lifts off the ground and before the swing leg makes contact with the ground, respectively. Similarly, let $v^+$ and $v^-$ represent the velocity of the CoM after the swing leg lifts off the ground and before the swing leg contacts the ground, respectively.  The state transitions ${\Delta _{\rm{SSP^-} \to \rm{SSP^+}}}$ in H-LIP dynamics are assumed to be smooth
\begin{gather}
	{\Delta _{\rm{SSP^-} \to \rm{SSP^+}}}:\left\{ \begin{array}{l}
		{v^ + } = {v^ - }\\
		{p^ + } = {p^ - }
	\end{array} \right. \label{eq:state_transitions}
\end{gather}
where the $+/-$ indicate the states after and before the transition, respectively.

Note that the SSP dynamics (\ref{eq:state_space_representation}) is a linear system. The locomotion during SSP can be straight to calculate. Let the SSP final state be $x^ - = [p^-,v^-]\t$ and the SSP initial state be $x^ +  = [p^+,v^+]\t$. The SSP final state $x^ - = [p^-,v^-]\t$ is calculated from the SSP initial state $x^ +  = [p^+,v^+]\t$ according to (\ref{eq:state_space_representation}) as
\begin{gather}
	x^ -  = {e^{{A_{\rm{SSP}}}{T_{\rm{SSP}}}}}x^ +  \label{eq:SSP_solution}
\end{gather}
where $T_{\rm{SSP}}$ is the swing period. The closed-form solution of the SSP dynamics are
\begin{align}
	\quad &\left\{ \begin{array}{l}
		p(t) = {c_1}{e^{\lambda t}} + {c_2}{e^{ - \lambda t}}\\
		v(t) = \lambda ({c_1}{e^{\lambda t}} - {c_2}{e^{ - \lambda t}})
	\end{array} \right.\label{eq:SSP}
\end{align} 
where $c_1 = \frac{1}{2e^{\lambda T_{\rm{SSP}}}}(p^{-} + \frac{1}{\lambda}v^{-})$ and $c_2 = \frac{1}{2e^{-\lambda T_{\rm{SSP}}}}(p^{-} - \frac{1}{\lambda}v^{-})$. Note that the H-LIP model is a passive LIP with no actuation. Equation (\ref{eq:SSP}) implies that the robot's CoM trajectory is given by $[p(t),v(t)]\t$, when the height of the CoM $z_{\rm{com}} \equiv z_0$.

To establish the robot's locomotion within a swing period for the subsequent gait design, we consider the SSP initial state (the start state of a step) after the previous contact and the SSP final state (the end state of a step) before the current contact as 
\begin{gather}
{\tilde{\Delta} _{\rm{SSP^+} \to \rm{SSP^-}}}:\left\{ \begin{array}{l}
	{v^ + } = {v^ - }\\
	{p^ + } = {p^ - } - L
\end{array} \right. \label{eq:H_LIPdynamics}
\end{gather}
where $L$ is the step size of the H-LIP. Once $T_{\rm{SSP}}$ is determined, the step length $L$ can be calculated according to (\ref{eq:SSP}) as
\begin{align}
	L = p(T_{\rm{SSP}}) - p(0) = T_1 p^- + T_2v^-\label{eq:u}
\end{align}
where 
\begin{gather}
T_1 = (1 - \frac{{{e^{\lambda {T_{\rm{SSP}}}}} + {e^{ - \lambda {T_{\rm{SSP}}}}}}}{2}),T_2 = \frac{1}{{2\lambda }}({e^{\lambda {T_{\rm{SSP}}}}} - {e^{ - \lambda {T_{\rm{SSP}}}}}).\nonumber
\end{gather}
Equation (\ref{eq:u}) illustrates the relationship between the step length and the final state of the SSP dynamics at the contact time. In other words, suppose that the CoM height remains constant at $z_{\rm{com}} \equiv z_0$, and that the dynamics described in (\ref{eq:H_LIPdynamics}) hold between the start and end states of a step, with $L$ given in (\ref{eq:u}). Under these conditions, the robot's dynamics follow H-LIP dynamics, which are composed of (\ref{eq:state_space_representation}) and (\ref{eq:state_transitions}). Based on these assumptions, the following gait planner design is proposed.

\section{Gait Planner Design}\label{sec:gait_design}
In this section, we propose a method to design a class of gait satisfying the H-LIP dynamics. To ensure that the robot satisfies the H-LIP dynamics, the relationship between the start and end states of a step, as described in (\ref{eq:H_LIPdynamics}), must be maintained, and the CoM height must remain constant at $z_{\rm{com}} = z_0$. We approximate the $X$-model and $Y$-model as the H-LIP model to design the robot's gait.


\subsection{Gait Design of X-model} \label{sec:motion_planner_x_model}

We will seek a class of joint trajectories parameterized by Bézier polynomials that satisfy the H-LIP dynamics and the contact condition based on kinematics of $X$-model. Due to the small mass of the foot link, we take the foot link and the shank as the whole link with mass $m_1$ to calculate the position and velocity of the CoM. Denote the mass of the thigh as $m_2$ and the torso as $m_3$. Let $l^x_1$ and $l^x_2$ be lengths of the shank and the thigh, respectively.

To adopt the SSP dynamics (\ref{eq:SSP_dynamics}) as the dynamics analysis for $X$-model, the $X$-model first should be ensured to approximated as the H-LIP model, that is, the height of CoM is constant approximately.
The CoM position $(x_{\rm{com}}, z^x_{\rm{com}})$ in $X$-model can be directly calculated as
\begin{align}
	x_{\rm{com}}(q^x(t)) =& \frac{m_1h^x_1 + m_2h^x_2 + m_2h^x_3 + m_1h^x_4+m_3h^x_5}{2m_1+2m_2+m_3}\label{eq:xCOMPosition}\\
	z^x_{\rm{com}}(q^x(t)) =& \frac{m_1v^x_1 + m_2v^x_2 + m_2v^x_3 + m_1v^x_4 + m_3v^x_5}{2m_1+2m_2+m_3}\label{eq:zCOMPosition}
\end{align}
where $h^x_i$ and $v^x_i$ is given in Appendix. The CoM velocity ($\dot{x}_{\rm{com}}(q^x,\dot{q}^x)$, $\dot{z}^x_{\rm{com}}(q^x,\dot{q}^x)$) is related to the position $q^x(t)$ and velocity $\dot{q}^x(t)$ and  can be expressed the form as
\begin{gather}
	\left[ {\begin{array}{*{20}{c}}
			{{{\dot x}_{{\rm{com}}}}}\\
			{{{\dot z^x}_{{\rm{com}}}}}
	\end{array}} \right] = {K_x}({q^x}){\dot q^x} \label{eq:form_xcom}
\end{gather}
where $K_x(q^x) \in \mathbb{R}^{2 \times 5}$ can be straightforwardly calculated based on (\ref{eq:xCOMPosition}) and (\ref{eq:zCOMPosition}).

When the height of the swing leg above the ground is zero, the swing leg contact the ground and the impact occurs. The occurrence of the impact event can be
denoted as
\begin{align}
\Gamma^x(q^x(t)) =& l^x_1 (\cos(q^x_1(t) + q^x_2(t)) - \cos(q^x_3(t) + q^x_4(t))) \nonumber\\ &+ l^x_2(\cos(q^x_2(t)) - \cos(q^x_3(t))).\label{eq:Gammax}
\end{align}

Note that when the robots is considered as H-LIP, the relationship between step length and the contact final state (\ref{eq:u}) should be satisfied. The swing leg contact the ground at time $t = 0$ and $t = T_{\rm{SSP}}$. Thus, the step length $L^x_q$ for $X$-model is given by
\begin{gather}
L^x_q = \frac{1}{2}H^x(q^x(T_{\rm{SSP}})) - H^x(q^x(0)) \label{eq:L}
\end{gather}
where 
\begin{align}
	H^x(t)=& l_1 (\sin(q^x_1(t) + q^x_2(t)) - \sin(q^x_3(t) + q^x_4(t))) \nonumber\\ &+ l_2(\sin(q^x_2(t)) - \sin(q^x_3(t))).\nonumber
\end{align}

Next, we present the following theorem for the desired joint trajectories of the $X$-model.  

\bthm\label{con:condition2.1}
Let $T_{\rm{SSP}} > 0$ be defined in (\ref{eq:SSP_solution}). Define the matrix
\begin{gather}
R = \left[ {\begin{array}{*{20}{c}}
		0&0&0&1&0\\
		0&0&1&0&0\\
		0&1&0&0&0\\
		1&0&0&0&0\\
		0&0&0&0&1
\end{array}} \right]\nonumber
\end{gather} representing the swap of the swing and stance legs and relabeling of the states.
A continuous trajectory $\phi^x(t): \mathbb{R} \to \mathbb{R}^5$ is the desired joint trajectory for the X-model if $(\phi^{x}(t), \dot{\phi}^x(t))$ satisfies the following properties
\begin{enumerate}
\item $\phi^{x}(0) = R \phi^{x}(T_{\rm{SSP}})$ and $\dot{\phi}^{x}(0) = R \dot{\phi}^{x}(T_{\rm{SSP}})$.
\item  $\Gamma^x(\phi^{x}(0)) = 0$ and $\Gamma^x(\phi^{x}(T_{\rm{SSP}})) = 0$ where $\Gamma^x(\bullet)$ is given in (\ref{eq:Gammax}).
\item $\Gamma^x(\phi^{x}(t)) > 0$, $\forall t \in(0, T_{\rm{SSP}})$.
\item ${K_x}({\phi^x}(T_{\rm{SSP}})){{\dot \phi }^x}(T_{\rm{SSP}}) = \left[ {\begin{array}{*{20}{c}}
		\frac{L_\phi ^x - {T_1}{x_{{\rm{com}}}}({\phi^x}({T_{{\rm{SSP}}}}))}{T_2}\\
		0
\end{array}} \right]$.
\item  $z^x_{\rm{com}}(\phi_x(t)) = z_0$, $\forall t \in(0, T_{\rm{SSP}})$ where $z_0$ is a constant.
\end{enumerate}
\ethm
Let us explain the motivation of the condition.  Suppose that the joint angles $q^x(t)$ starts to follow the desired joint trajectory $\phi^x(t)$.
The first term of property 1) ensures that the impact occurs at time $T_{\rm{SSP}}$, at which point the swing leg switches to the stance leg, and its joint angle remains on the trajectory. The second term of property 1) guarantees that the first equation in (\ref{eq:H_LIPdynamics}) holds.
Property 2) indicates that the swing leg will touch the ground at  $t = T_{\rm{SSP}}$.  Property 3) ensures that the swinging leg remains above the ground for $t \in (0, T_{\rm{SSP}})$. Property 4) guarantees that the end state of a step satisfies the second equation in (\ref{eq:H_LIPdynamics}). Property 5) ensures that the height of the CoM remains constant. The second term of property 1), along with properties 4) and 5), ensures that the dynamics of the $X$-model are approximated as H-LIP dynamics, while the first term of property 1), along with properties 2) and 3), ensures that the $X$-model satisfies the contact condition.

Bézier polynomials have been widely employed in previous studies to parameterize joint trajectories, particularly due to their numerical stability during optimization \cite{westervelt2003hybrid}. The parameters of these polynomials can be efficiently determined through nonlinear optimization techniques. By applying the given theorem \ref{con:condition2.1}, the nonlinear optimization method can be used to obtain the desired trajectory $\phi^{x}(t)$. However, these methods do not meet the real-time requirements. Next, we aim to propose a real-time and highly effective method to obtain the desired trajectory $\phi^{x}(t)$.

We first parameterize the trajectory $\phi^{x}(t) = [\phi_i^x(t)] \in \mathbb{R}^{5 \times 1}$ for the generalized coordinates $q^x_i$ using Bézier polynomials of order $N$ as 
\begin{gather}
	\phi ^x_i(t) = \sum\limits_{j = 0}^{N}  {\left( {\begin{array}{*{20}{c}}
				N\\
				j
		\end{array}} \right)} {\alpha^i _{j}}{(1 - \frac{t}{T_{\rm{SSP}}})^{N - j}}{(\frac{t}{T_{\rm{SSP}}})^j}\label{eq:phit}
\end{gather}
where $\alpha^i_j$ for $i = 1,\cdots, 5$ and $j = 1,\cdots, N$ are the parameter of the Bézier polynomials and $T_{\rm{SSP}}$ is defined in (\ref{eq:SSP_solution}).

To make the parameterized trajectory tractable and inspired by the
observation of trajectory for a planar biped gait design in \cite{grizzle2014models, li2023stable}, we further adopt a particular choice as
\begin{gather}
	\phi_1^{x}(T_{\rm{SSP}}) = \phi_4^{x}(T_{\rm{SSP}}). \label{eq:phi_1=phi_4}
\end{gather}
Equality (\ref{eq:phi_1=phi_4}) means that the angle between the thigh and the stance leg is equal to the angle between the thigh and the shank leg when the contact occurs.

The step length of the walking behavior is a design variable. The following lemma provides the relationship between the step length and the parameter of the trajectory (\ref{eq:phit}).

\blem\label{lem:3.1}
Consider the trajectory (\ref{eq:phit}) with the condition (\ref{eq:phi_1=phi_4}). Suppose that $\alpha^1_1 = \alpha^4_N$, $\alpha^2_1 = \alpha^3_N$, $\alpha^3_1 = \alpha^2_N$, $\alpha^4_1 = \alpha^1_N$, and
\begin{gather}
	l^x_1\sin(\alpha^1_{1} + ({\alpha^2_{1} + \alpha^3_{1}})/{2}) = -l^x_2\sin(({\alpha^2_{1}+\alpha^3_{1}})/{2}) \label{eq:relationship_l1_l2}
\end{gather}
If the robot track the trajectory $(\phi^x(t), \dot{\phi}^x(t))$, the step length $L^x_{\phi} $ of the robot is
\begin{gather}
L ^x_{\phi} =  - \frac{{2{l^x_1}\sin ({\alpha^1 _{{N}}})\sin (({{\alpha^2 _{{N}}} - {\alpha^3 _{{N}}}})/{2})}}{{\sin ({({{\alpha^2 _{{N}}} + {\alpha^3 _{{N}}}})}/{2})}}.\label{eq:stepLength}
\end{gather}
\elem
\proofnow
From the condition (\ref{eq:phi_1=phi_4}), we have
$
\alpha^1_N = \alpha^4_N.
$ Since $\alpha^1_1 = \alpha^4_N$ and $\alpha^4_1 = \alpha^1_N$ according to the condition in lemma \ref{lem:3.1}, it gives $\alpha^1_1 = \alpha^4_1 = \alpha^1_N = \alpha^4_N$.
From (\ref{eq:L}), the step length is calculated as
\begin{align}
	{L^x_{\phi}} =& 2{l^x_1}\sin (\frac{{\alpha _N^2 - \alpha _N^3}}{2})\cos (\alpha _N^1 + \frac{{\alpha _N^2 + \alpha _N^3}}{2}) \nonumber\\&+ 2{l^x_2}\sin (\frac{{\alpha _N^2 - \alpha _N^3}}{2})\cos (\frac{{\alpha _N^2 + \alpha _N^3}}{2}). \label{eq:Lx}
\end{align}
By using $\alpha^1_1 = \alpha^4_1 = \alpha^1_N = \alpha^4_N$, $\alpha^2_1 = \alpha^3_N$, and $\alpha^3_1 = \alpha^2_N$, substituting (\ref{eq:relationship_l1_l2}) into (\ref{eq:Lx}) gives (\ref{eq:stepLength}).
\eproof

If $\phi^{x}(t)$ are further specified in (\ref{eq:phit}) and the condition is selected in (\ref{eq:phi_1=phi_4}), we can obtain
corollary \ref{cor:X-model}.

\bcorollary\label{cor:X-model}
Let $e_1 = [1,1,1,1, 0]\t, e_2 = [0,1,1,0,0]\t, e_3 = [1,1,-1,-1,0]\t$, and $e_4 = [0,1,-1,0,0]\t$. Define the vectors
\begin{gather}
{\alpha}_{j} = [\alpha^1_{j}, \alpha^2_j, \alpha^3_j, \alpha^4_j, \alpha^5_j]\t.\nonumber
\end{gather}
Suppose the condition in Lemma \ref{lem:3.1} is satisfied. The trajectory $\phi^{x}(t)$ in (\ref{eq:phit}) is the desired joint trajectory for $X$-model if the trajectory $(\phi^{x}(t), \dot{\phi}^x(t)$ with parameter $\alpha^i_{j}$ for $ i = 1,\cdots 5$ and $j = 1,\cdots, N$ satisfies the following conditions:
\begin{enumerate}
\item $\alpha _2^1 = \alpha _N^4 + \alpha _1^1 - \alpha _{N - 1}^4,	\alpha _2^2 = \alpha _N^3 + \alpha _1^2 - \alpha _{N - 1}^3,\\ \alpha _2^3 = \alpha _N^2 + \alpha _1^3 - \alpha _{N - 1}^2,	\alpha _2^4 = \alpha _N^1 + \alpha _1^4 - \alpha _{N - 1}^1$.
\item $l^x_1\sin(e\t_1\phi^{x}(t)/2)\sin(e\t_3\phi^{x}(t)/2) + l^x_2\sin(e\t_2\phi^{x}(t)/2)\sin(e\t_4\phi^{x}(t)/2) < 0$, $\forall \in (0, T_{\rm{SSP}})$.
\item 
${K_x}({\alpha _N})({\alpha _N} - {\alpha _{N - 1}}) =\frac{{{T_{SSP}}}}{{N{T_2}}} \left[ {\begin{array}{*{20}{c}}
		{L_\phi ^x - {T_1}{x_{{\rm{com}}}}({\alpha _N})}\\
		0
\end{array}} \right]$.
\item $z^x_{\rm{com}}(\phi_x(t)) = z_0$, $\forall t \in(0, T_{\rm{SSP}})$ where $z_0$ is a constant.
\end{enumerate}
\ecorollary

\proofnow
We give the proof according to the properties of theorem \ref{con:condition2.1}. 
By the condition in lemma \ref{lem:3.1}, it gives
$
\alpha^1_1 = \alpha^4_N, \alpha^2_1 = \alpha^3_N 
$. By the property 4) in corollary \ref{cor:X-model}, we have
$
\alpha^5_1 = \alpha^5_N
$. It follows that
$
\phi^x(0) = R\phi^x(T_{\rm{SSP}})
$. Note that $\dot{\phi}_i^x(0) = \frac{N}{T_{\rm{SSP}}}(\alpha^i_1 - \alpha^i_0)$ and $\dot{\phi}_i^x(T_{\rm{SSP}}) = \frac{N}{T_{\rm{SSP}}}(\alpha^i_N - \alpha^i_{N-1})$. By the property 1) in corollary \ref{cor:X-model}, we have $\dot{\phi}^x(0) = R\dot{\phi}^x(T_{\rm{SSP}})$. Then, the property 1) in theorem \ref{con:condition2.1} is satisfied. From (\ref{eq:relationship_l1_l2}), we have
\begin{align}
- 2{l^x_1}\sin (\alpha _1^1 + \frac{{\alpha _1^2 + \alpha _1^3}}{2})\sin (\frac{{\alpha _1^2 - \alpha _1^3}}{2}) &- 2{l^x_2}\sin (\frac{{\alpha _1^2 + \alpha _1^3}}{2})\nonumber\\ \times &\sin (\frac{{\alpha _1^2 - \alpha _1^3}}{2}) = 0.	
\end{align} 
It follows that
$
{\Gamma ^x}(\phi^x(0)) = 0
$. By the property 1) in theorem \ref{con:condition2.1} and $
{\Gamma ^x}(\phi^x(0)) = 0
$, it leads to property 2) in theorem \ref{con:condition2.1}. By property 2) in corollary \ref{cor:X-model}, for $t \in (0,T_{\rm{SSP}})$, it gives
\begin{align}
- &2{l^x_1}\sin (\phi^x_1(t) + \frac{{\phi^x_2(t) + \phi^x_3(t)}}{2})\sin (\frac{{\phi^x_2(t) - \phi^x_3(t)}}{2}) \nonumber\\&- 2{l^x_2}\sin (\frac{{\phi^x_2(t) + \alpha^x_3(t)}}{2})\sin (\frac{{\phi^x_2(t) - \phi^x_3(t)}}{2}) > 0.	
\end{align}
That is, $\Gamma^x(\phi^x(t)) > 0$, $\forall t \in (0,T_{\rm{SSP}})$, the property 3) in theorem \ref{con:condition2.1} is satisfied. Since $\dot{\phi}_i^x(T_{\rm{SSP}}) = \frac{N}{T_{\rm{SSP}}}(\alpha^i_N - \alpha^i_{N-1})$, the property 3) in corollary \ref{cor:X-model} leads to property 4) in theorem \ref{con:condition2.1}. The property 4) in corollary \ref{cor:X-model} is the same as the property 5) in theorem \ref{con:condition2.1}. All of the properties in theorem \ref{con:condition2.1} are satisfied, the trajectory $\phi^x(t)$ in (\ref{eq:phit}) is the desired joint trajectory for X-model.
\eproof

With corollary \ref{cor:X-model}, the trajectory $\phi^x(t)$ with parameter $\alpha^i_j$ for $i = 1,\cdots,5$ and $j = 1,\cdots,N$ can be determined by using the algorithm \ref{alg:search}.

\IncMargin{1em}
\begin{algorithm}[htbp]
	\While{Step length $L^x_{\phi}$ and walking speed $V_x$ are given}{
	Let $T_{\rm{SSP}} = L^x_{\phi}/V_x$.\\
	
	Choose $\alpha^1_1$, $\alpha^1_j$, $\alpha^2_j$, $\alpha^3_j$, $\alpha^4_j$ for $j = 3,\cdots,N-2$ and  ${\alpha}^k_{N-1}$ for $k = 1,2,3$.\\
		Calculate $({\alpha^2_1 + \alpha^3_1})/{2}$ according to (\ref{eq:relationship_l1_l2}) and $\alpha^1_1$.\\
		Let $\alpha^1_N = \alpha^1_1$, $ \alpha^4_1 = \alpha^1_1$. \\
		Calculate $\alpha^2_1$ and $\alpha^3_1$ using $({\alpha^2_1 + \alpha^3_1})/{2}$, $L^x_{\phi}$ in (\ref{eq:stepLength}), and property 1) of corollary \ref{con:condition2.1}.\\
		Calculate $\alpha^5_1$ according to $z^x(\phi_{x}(0)) = z_0$ and $\alpha^i_1$ for $i = 1,2,3,4$.\\
		Let $\alpha^2_N = \alpha^3_1$, $\alpha^3_N = \alpha^2_1$, $\alpha^4_N = \alpha^1_1$, and $\alpha^5_N = \alpha^5_1$.\\
        Calculate ${\alpha}^4_{N-1}$ and ${\alpha}^5_{N-1}$ according to property 4) in corollary \ref{con:condition2.1}, ${\alpha}^k_{N-1}$ for $k = 1,2,3$, and $\alpha_N = [\alpha^1_N, \alpha^2_N, \alpha^3_N, \alpha^4_N, \alpha^5_N]\t$.\\
        Calculate $\alpha^1_2$, $\alpha^2_2$, $\alpha^3_2$, and $\alpha^4_2$ according to property 2) in corollary \ref{con:condition2.1}.
	
		\eIf{Property 4) in corollary \ref{cor:X-model} is satisfied}{Go to step 15.}{Go to step 3.}

	Calculate $\phi^x_5(t)$ according to $z^x_{\rm{com}}(\phi_{x}(t)) = z_0$.\\
	Calculate $\alpha^5_j$ according to $\phi^x_5(t)$.
	}
\caption{Gait Design of X-model}
\label{alg:search}
\end{algorithm}
\brem
To ensure real-time gait design using Algorithm \ref{alg:search}, suitable initial values for $\alpha^{i}_{j}$ must be provided. These parameters are then explored through uniform sampling in the vicinity of the initial values. Additionally, the parameter $\alpha^{i}_{j}$ in Step 3 of algorithm \ref{alg:search} can be determined via offline nonlinear optimization.
\erem

\subsection{Gait Design of Y-model}
\begin{figure}[!ht]
	\centering
	\includegraphics[scale=0.42]{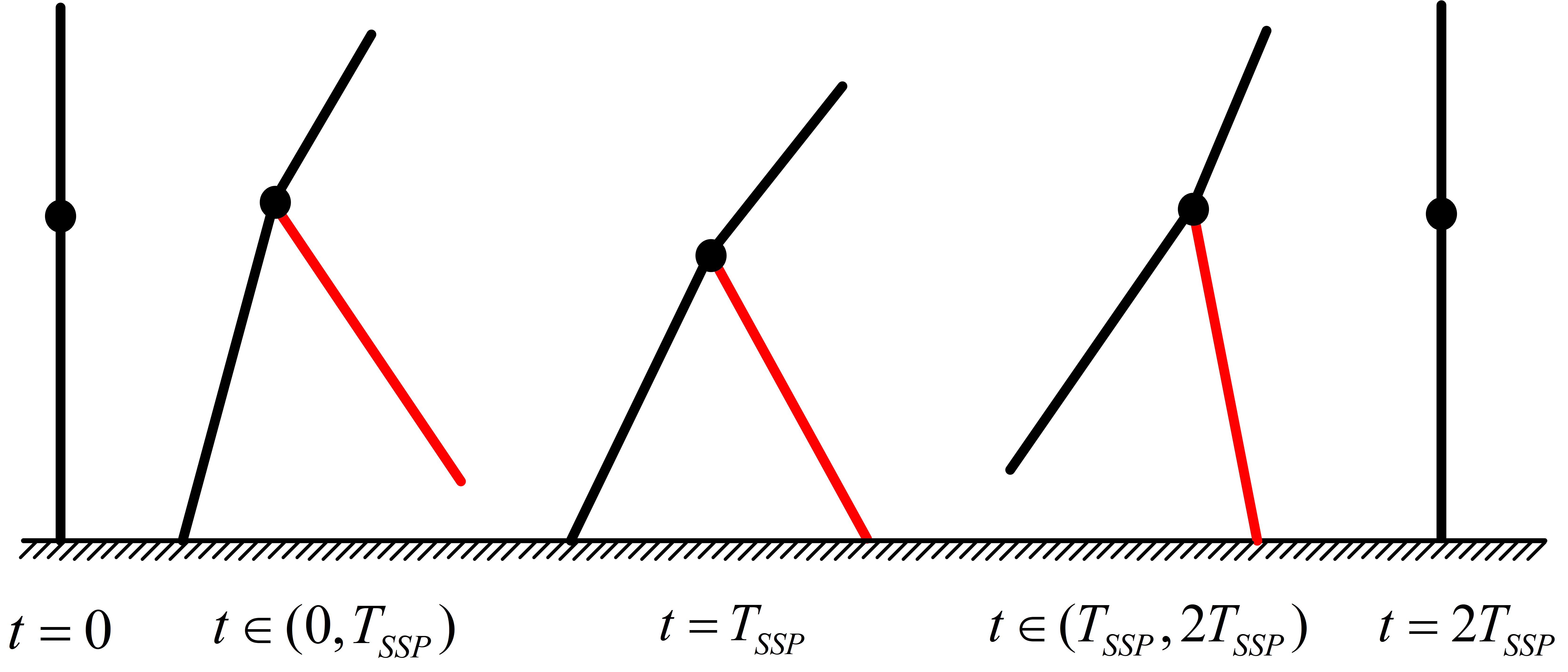}
	\caption{Different phases for $Y$-model.}
	\label{fig_gait_design_for_Y}
\end{figure}

\begin{figure}[!ht]
	\centering
	\includegraphics[scale=0.52]{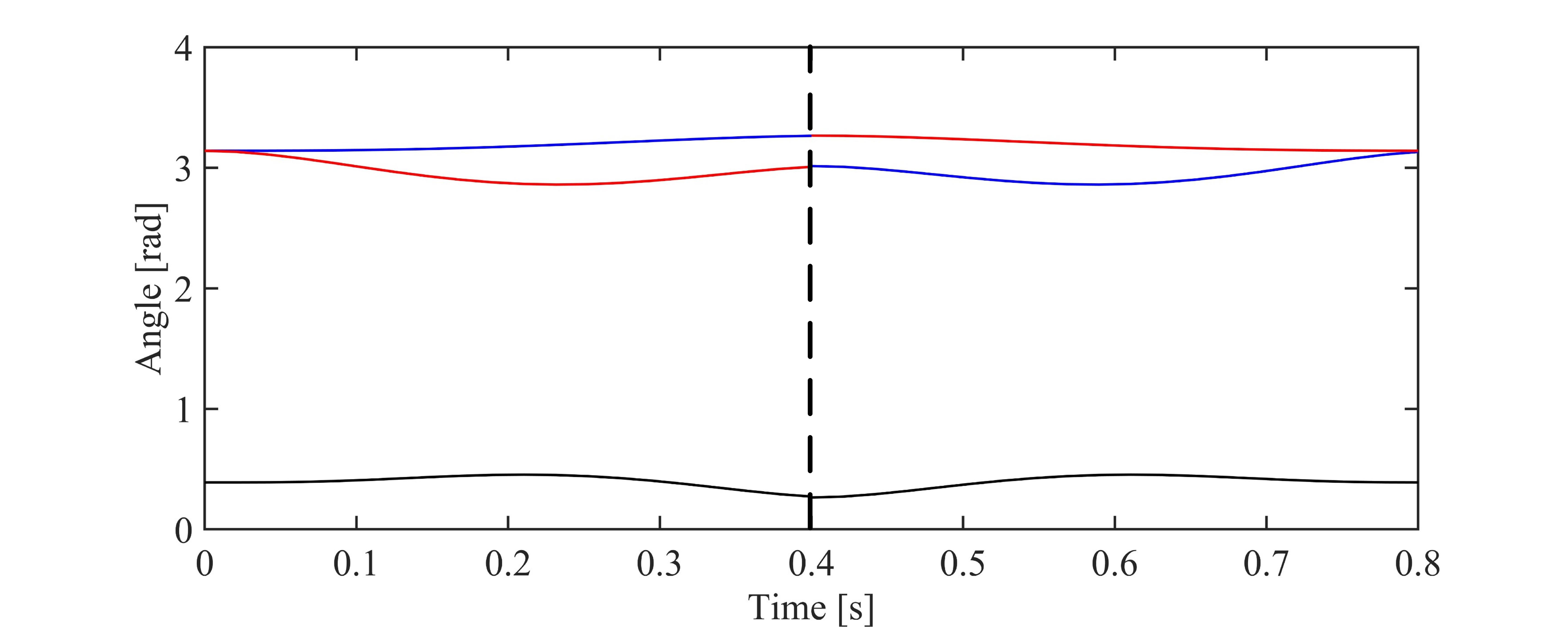}
	\caption{The desired trajectory corresponding to different phases of the $Y$-model (blue: $q^y_1(t)$, red: $q^y_2(t)$, black: $q^y_3(t)$, $T_{\rm{SSP}} = 0.4$)}
	\label{fig_gait_design_for_Y_des}
\end{figure}

 Unlike the gait design of the $X$-model, the $Y$-model requires two swing periods ($2T_{\rm{SSP}}$) to walk along the $Y$-axis direction with a step length $L_y$ as the physical legs cannot collide, as shown in Fig. \ref{fig_gait_design_for_Y}. The corresponding joint trajectory for $q^y(t)$ is shown in Fig. \ref{fig_gait_design_for_Y_des}.

Let $l^y_1$ and $l^y_2$ represent the lengths of the stance leg and swing leg in the $Y$-model, respectively. The lengths of these legs are influenced by the motion of the $X$-model. For simplicity, we assume that the leg lengths in the $Y$-model correspond to those in the $X$-model at time $t = 0$ (height of the torso is equivalent for $X$-model and $Y$-model in Fig. \ref{fig_decouple_model}). Then, the lengths of the stance leg and swing leg can be calculated as
\begin{gather}
	{l}^y_1 = \sqrt{(l^x_1)^2 + (l^x_2)^2 - 2l^x_1l^x_2\cos(\pi - q^x_1(0))}\nonumber\\
	{l}^y_2 = \sqrt{(l^x_1)^2 + (l^x_2)^2 - 2l^x_1l^x_2\cos(\pi - q^x_4(0))}. 
\end{gather}
The CoM position $(y_{\rm{com}}, z^y_{\rm{com}})$ in Y-model is
\begin{gather}
	y_{\rm{com}} = \frac{(m_1+m_2)h^y_1 + (m_1+m_2)h^y_2+m_3h^y_3}{2m_1+2m_2+m_3}\nonumber\\
	z^y_{\rm{com}} = \frac{(m_1+m_2)v^y_1 + (m_1+m_2)v^y_2+m_3v^y_3}{2m_1+2m_2+m_3}
\end{gather}
where $h^y_i$ and $v^y_i$ are given in Appendix. Similar to (\ref{eq:form_xcom}), the CoM velocity of the H-LIP model in $Y$-model can be expressed as
\begin{gather}
\left[ {\begin{array}{*{20}{c}}
		{{{\dot y}_{\rm{com}}}}\\
		{{{\dot z^y}_{\rm{com}}}}
\end{array}} \right] = {K_y}({q^y}){\dot q^y}.
\end{gather}
The occurrence of the impact event for $Y$-model is
\begin{gather}
\Gamma^y(q^y(t)) = l^y_1\cos(\pi - q^y_1) - l^y_2\cos(q^y_2 - \pi) \label{eq:Gammay}
\end{gather}
and the step length of $Y$-model is given as
\begin{gather}
L^y_q = -2l_1^y \sin(\frac{q^y_2(T_{\rm{SSP}}) - q^y_1(T_{\rm{SSP}})}{2}). \label{eq:Ly}
\end{gather}

Furthermore, we outline the following conditions for the desired joint trajectories of the $Y$-model.

\bthm\label{condition3.2}
Let $T_{\rm{SSP}} > 0$ be defined in (\ref{eq:SSP_solution}). A continuous trajectory $\phi^{y}(t) = [\phi^{y}_i(t)] : \mathbb{R} \to \mathbb{R}^3$ is the desired joint trajectory for $Y$-model if $(\phi^{y}(t), \dot{\phi}^y(t))$ satisfies the following properties
\begin{enumerate}
\item ${\phi}^{y}_1(0) = {\phi}^y_2(0)$ and $\dot{\phi}^{y}_1(T_{\rm{SSP}}) = \dot{\phi}^y_2(T_{\rm{SSP}})$.
\item $\Gamma^y(\phi_{y}(0)) = 0$ and $\Gamma^y(\phi_{y}(T_{\rm{SSP}})) = 0$.
\item $\Gamma^y(\phi_{y}(t)) > 0$, $\forall t \in(0, T_{\rm{SSP}})$.
\item ${K_y}({\phi^y}(T_{\rm{SSP}})){{\dot \phi }^y}(T_{\rm{SSP}}) = \frac{1}{{2T_2}}\left[ {\begin{array}{*{20}{c}}
		{L_\phi ^y - {T_1}{y_{{\rm{com}}}}({\phi^y}({T_{{\rm{SSP}}}}))}\\
		0
\end{array}} \right]$.
\item $z^y_{\rm{com}}(\phi^y(t)) = z_0$, $\forall t \in(0, T_{\rm{SSP}})$ where $z_0$ is a constant.
\end{enumerate}
\ethm

We use the following Bézier polynomials to parameterize the trajectory $\phi^{y}(t) = [\phi_i^{y}(t)] \in \mathbb{R}^3$ as
\begin{gather}
	\phi^y_i(t) = \sum\limits_{j = 0}^{M}  {\left( {\begin{array}{*{20}{c}}
				M\\
				j
		\end{array}} \right)} {\beta^i _{j}}{(1 - \frac{t}{T_{\rm{SSP}}})^{M - j}}{(\frac{t}{T_{\rm{SSP}}})^j}.\label{eq:phity}
\end{gather}

Using the polynomials in (\ref{eq:phity}) to parameterize $\phi^y(t)$, the following lemma provides the relationship between the step length of the $Y$-model and the polynomial parameters.

\blem \label{lem3.2}
Consider the trajectory (\ref{eq:phity}). Suppose the parameter $\beta^1_M$ and $\beta^2_M$ satisfies
\begin{gather}
	\beta^1_M + \beta^2_M = 2\pi. \label{eq:beta_condition}
\end{gather}
If the $Y$-model track the trajectory $(\phi^y(t),\dot{\phi}^y(t))$, the step length $L^y_{\phi}$ of the $Y$-model is
\begin{gather}
L_\phi ^y = -2l_1^y\sin (\frac{\beta _M^2 - \beta_M^1}{2}). \label{eq:L_phi}
\end{gather}
\elem

Using the parameters in trajectory (\ref{eq:L_phi}), the following corollary can be derived from Theorem \ref{lem3.2}.
\bcorollary\label{cor:corallary2.2}
Define the vectors 
\begin{gather}
	\beta_j = [\beta^1_j,\beta^2_j,\beta^3_j,\beta^4_j,\beta^5_j]\t.\nonumber
\end{gather}
The trajectory $\phi^{y}(t)$ in (\ref{eq:phity}) is the desired joint trajectory for Y-model if the trajectory $(\phi^{y}(t),\dot{\phi}^y(t))$ with parameter $\beta^i_{j}$ satisfies the following conditions:
\begin{enumerate}
\item $\beta^1_1 = \beta^2_1 = \pi$ and $\beta^2_{M-1} = -\beta^1_M + \beta^1_{M-1}  + \beta^2_M $.
\item $\sin(({\phi^y_2(t)-\phi^y_1(t)})/{2})\cos(({\phi^y_2(t)+\phi^y_1(t)})/{2}) > 0$, $\forall t \in (0, T_{\rm{SSP}})$.
\item ${K_y}({\beta _M})({\beta _M} - {\beta _{M - 1}}) =\frac{{{T_{\rm{SSP}}}}}{{2M{T_2}}} \left[ {\begin{array}{*{20}{c}}
		{L_\phi ^y - {T_1}{y_{{\rm{com}}}}({\beta _M})}\\
		0
\end{array}} \right]$.
\item $z^y_{\rm{com}}(\phi^y(t)) = z_0$, $\forall t \in(0, T_{\rm{SSP}})$ where $z_0$ is given in property 4) in corollary \ref{cor:X-model}.
\end{enumerate}
\ecorollary

\proofnow
We give the proof according to the properties of theorem \ref{condition3.2}. 
According to property 1) in corollary \ref{cor:corallary2.2}, it leads to the property 1) in theorem \ref{condition3.2}. Substituting the property 1) in corollary \ref{cor:corallary2.2} into (\ref{eq:Gammay}) gives the property 2) in theorem \ref{condition3.2}. From the property 2) in corollary \ref{cor:corallary2.2}, it gives
\begin{gather}
l^y_1\cos(\pi - \phi^y_1(t)) - l^y_2\cos(\phi^y_2(t) - \pi) > 0, \forall t \in (0,T_{\rm{SSP}}).\nonumber
\end{gather}   
It follows that the property 3) in corollary \ref{cor:corallary2.2} is satisfied. Since $\dot{\phi}^y(T_{\rm{SSP}}) = \frac{M}{T_{\rm{SSP}}}(\beta_M - \beta_{M-1})$, the property 3) leads to the property 4) in theorem \ref{condition3.2}. Note that the property 4) in corollary \ref{cor:corallary2.2} is the same as the property 5) in theorem \ref{condition3.2}. All of the properties in theorem \ref{cor:corallary2.2} are satisfied, the trajectory $\phi^y(t)$ is the desired joint trajectory for $Y$-model.
\eproof

Similar to algorithm 1, we propose the following algorithm for designing the gait of the $Y$-model.

\IncMargin{1em}
\begin{algorithm}[htbp]
	\While{Walking velocity $V_y$ is given}{
	Let $L^y_{\phi} = T_{\rm{SSP}}V_y$.\\
	Calculate ${\beta^1_M}$ according to $L^y_{\phi}$ in (\ref{eq:L_phi}).\\
	Calculate $\beta^2_M$ according to (\ref{eq:beta_condition}).\\
	Calculate $\beta^3_M$ according to using $z^y_{\rm{com}}(\beta_M) = z_0$ in property 4) of corollary \ref{cor:corallary2.2}, $\beta^1_M$, and $\beta^2_M$.\\
	Let $\beta^1_1 = \pi$ and $\beta^2_1 = \pi$.\\
	Calculate $\beta^1_{M-1}$, $\beta^2_{M-1}$, and $\beta^3_{M-1}$ according to property 1) and property 3) in corollary \ref{cor:corallary2.2} with $\beta_M = [\beta^1_M,\beta^2_M, \beta^3_M]\t$.\\
	Choose $\beta^1_j$, $\beta^2_j$ for $j = 2,3,\cdots, M-2$.\\
	\eIf{Property 2) in corollary \ref{cor:corallary2.2} is satisfied}{
	Go to step 8.
	}{
	Go to step 13.
	}
	Calculate $\phi^y_3(t)$ according to property 4) in corollary \ref{cor:corallary2.2}, $\phi^y_1(t)$, and $\phi^y_2(t)$.\\
	Calculate $\beta^3_j$ for $j = 1,\cdots,M-2$ according to $\phi^y_3(t)$.
}
\caption{Gait Design of $Y$-model}
\label{alg:search1}
\end{algorithm}

The joint trajectories for the $X$-model and $Y$-model can be obtained using algorithms \ref{alg:search} and \ref{alg:search1}. As established by the kinematic relationships in Table \ref{table:relationship_Decopule_model}, the robot's joint trajectories can be readily obtained.

\begin{table}[!h]
	\begin{center}
		\caption{Kinematics transformation between the $X/Y$-model and the robot}
		\begin{tabular}{c|c}
			\hline
			Angle in Real Humanoid Robot & 	Angle in Decouple Model\\
			\hline
			$q_{\rm{L-aky}}$ & $-q^x_3-q^x_4$\\
			\hline
			$q_{\rm{L-kny}}$ & $-q^x_4$\\
			\hline
			$q_{\rm{L-hpy}}$ & $q^x_3 + q^x_5$\\
			\hline
			$q_{\rm{R-aky}}$ & $-q^x_1-q^x_2$\\
			\hline
			$q_{\rm{R-kny}}$ & $-q^x_1$\\
			\hline
			$q_{\rm{R-hpy}}$ & $q^x_2 + q^x_5$\\
			\hline
			$q_{\rm{L-akx}}$ & $q^y_1$\\
			\hline
			$q_{\rm{L-hpx}}$ & $q^y_1 - q^y_3$\\
			\hline
			$q_{\rm{R-akx}}$ & $q^y_2$\\
			\hline
			$q_{\rm{R-hpx}}$ & $q^y_2 - q^y_3$\\
			\hline
			$q_{\rm{L-hpz}}$ & $0$\\
			\hline
			$q_{\rm{R-hpz}}$ & $0$\\
			\hline
		\end{tabular}\label{table:relationship_Decopule_model}
	\end{center}
\end{table}

\section{Learning Framework}\label{se:reward_composition}

\begin{figure*}[htbp]
	\centering
	\includegraphics[scale=0.76]{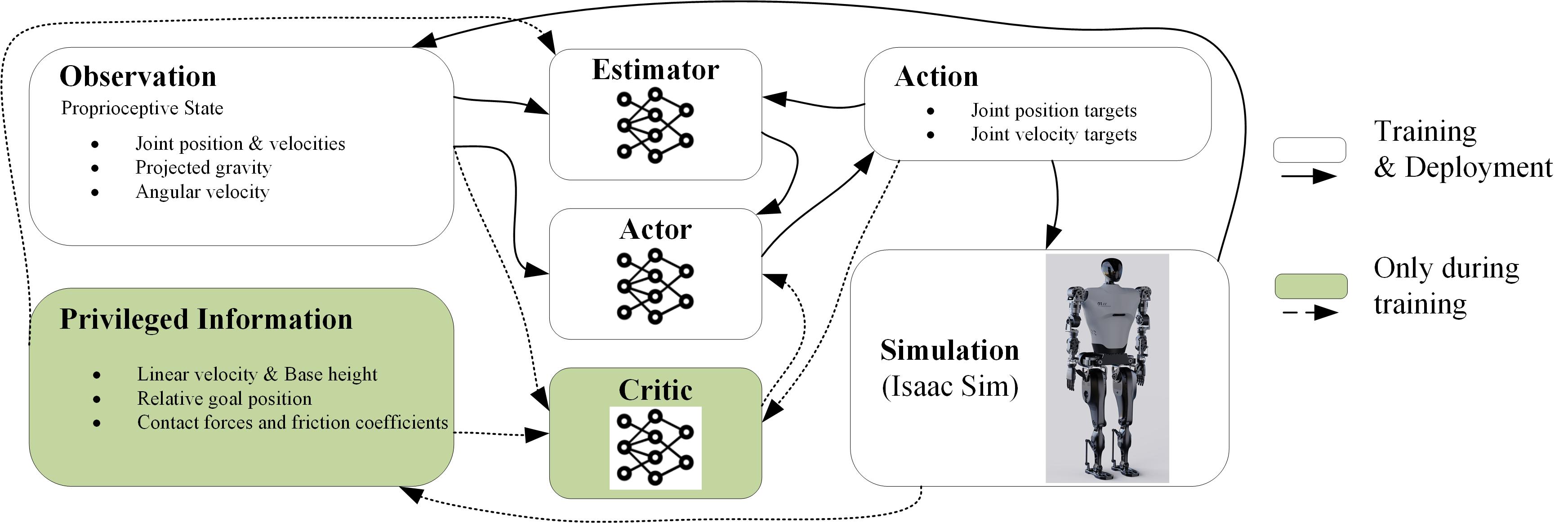}
	\caption{Overview of the learning framework.}
	\label{fig_framework}
\end{figure*}

By using reinforcement learning to achieve locomotion, and assuming access only to proprioceptive measurements from the inertial measurement unit and joint encoders, the dynamics of legged locomotion can be modeled as an infinite-horizon partially observable Markov decision process \cite{wang2024cts}. The overview of the learning framework is shown in Fig. \ref{fig_framework}. The reinforcement learning framework consists of three main sub-networks: the actor, the critic, and the estimator, with the learning process involving two components: optimizing the actor-critic and optimizing the estimator. In this study, the actor-critic is optimized using the Proximal Policy Optimization algorithm, and the estimator is optimized using regression techniques. The learning framework has been widely applied, and the details of the framework are presented in \cite{Chamorro2024ReinforcementLF}. The reward is a critical design element of the learning framework, influencing both the learning rate and locomotion performance. In \cite{9561814}, a periodic reward composition is proposed for learning bipedal gaits. In this section, we design three reward functions to establish a periodic bipedal gait. Inspired by \cite{9561814}, the first and second reward functions are designed to ensure periodic walking gaits for humanoid robots, while the third reward function is aimed at ensuring that the actual joint angles track the robot desired joint trajectory, as outlined in Section \ref{sec:gait_design}, thereby accelerating the learning rate.

In bipedal locomotion, each leg undergoes two characteristic phases: the stance phase and the swing phase. We define the first-term reward function $r^{\rm{vf}}_t$ by penalizing the foot's horizontal velocity during the stance phase and the foot forces during the swing phase. The first reward function is established as:
\begin{gather}
r^{\rm{vf}}_t = \sum\limits_{i = 1}^{{n_{\rm{leg}}}} {[{C_i}(1 - {e^{ - ||{v^{\rm{xy}}_{i}}||/1.25}}) + (1 - {C_i})(1 - {e^{ - ||{f^{\rm{xyz}}_i}||/50}})]}\nonumber
\end{gather} 
where $v^{\rm{xy}}_i$ represent the velocity along $x$- and $y$- axes of $i$-th leg, $f^{\rm{xyz}}_i$ represent the foot force of $i$-th leg, and $C_i$ represents the characteristic phase of the $i$-th leg. The value of $C_i$ is chosen as 
\begin{gather}
C_i(\tau'_i) = {sigmoid}(10\sin(2\pi\tau'_i)) \label{eq:Ci}
\end{gather}
with
\begin{gather}
{\tau '_i} = \left\{ \begin{array}{l}
	0.5{\tau _i}/{r_{\rm{st}}}, \quad \quad \quad \quad \quad \quad \quad \quad \quad \tau_i\in [0,r_{\rm{st}}]\\
	0.5 + 0.5({\tau _i} - {r_{\rm{st}}})/(1 - {r_{\rm{st}}}),\;\; \tau_i\in(r_{\rm{st}},1]
\end{array} \right.
\end{gather}
where $\tau_i = (f_c t + \Delta \tau_i)$ mod 1, with $f_c$ being the frequency of the gait cycle, represents the leg phase in the interval $[0,1]$, and $r_{\rm{st}}$ is the constant selected to represent the ratio of the time spent in stance phase relative to the entire gait period. For the periodically symmetric bipedal gait in this paper, $r_t$ is selected as 0.5.
In (\ref{eq:Ci}), $C_i$ is set to $(0.5,1]$ when the $i$-th leg is in the stance phase, and $C_i$ is set in $[0,0.5]$ when the $i$-th leg in the swing phase. The $sigmoid$ function is used to ensure a smooth transition of $C_i$ between 0 and 1, or vice versa. The $sigmoid$ function is selected as
$
sigmoid(x) = \frac{1}{1+e^{-x}}.
$

Define the second reward function $r^{\rm{cs}}_{t}$ to granting the $i$-th leg is in the correct phase, as described by the value of $C_i$. The second reward function is designed as:
\begin{gather}
r^{\rm{cs}}_{t} = \sum\limits_{i = 1}^{{n_{\rm{leg}}}} {[1 + 1.3(2{C_i}{S_i} - {C_i} - {S_i})]} 
\end{gather}
with 
\begin{gather}
{S_i} = \left\{ \begin{array}{l}
	1 \quad ||f^{\rm{xyz}}_i||>5\\
	0 \quad ||f^{\rm{xyz}}_i||\le 5
\end{array} \right.
\end{gather}
where $S_i = 1$ indicates that the $i$-th leg of the robot in the learning environment is in contact with the ground, and $S_i = 0$ indicates that the $i$-th leg is not in contact with the ground.

Define the third reward function $r_t^{\rm{track}}$ to ensure that the joint angles of robot in learning environment track the desired joint trajectory designed in section \ref{sec:gait_design}. Let $\rm{q}_{\rm{des}} \in \mathbb{R}^{12}$ represent the desired joint angles of the real robot, obtained by sampling the desired trajectory $\phi^x(t)$ or $\phi^y(t)$ and performing a coordinate transformation to obtain the desired joint angles, and $\rm{q}_{\rm{actual}} \in \mathbb{R}^{12}$ represent the actual joint angles of the robot. 
The third reward is designed  as
\begin{gather}
r_t^{\rm{track}} = {e^{ - 2||\rm{q}_{\rm{des}} - \rm{q}_{\rm{actual}}||}}.
\end{gather}

\section{Design Example and Performance Comparison}\label{sec:design_example} 
In this section, we first design the gait according to algorithms 1 and 2 presented in section \ref{sec:gait_design}, and then design reward functions within the reinforcement learning framework. We compare the training performance using the reward composition proposed in Section \ref{se:reward_composition} with that of a setup without the reward composition. The parameters of the studied humanoid robot are provided in Table \ref{table:parameter_of_humanoid_robot}. The distance between the CoM of the link and the corresponding joint is assumed to be at the center of the link. Specifically, $r^x_i = \frac{1}{2}l^x_i$ for $i = 1,2,3,4,5$, and $r^y_j = \frac{1}{2}l^y_j$ for $j = 1,2,3$. The gait design examples for both the X-model and Y-model are provided in section \ref{sec:gait_example}, while the simulation comparison is presented in section \ref{sec:performance_comparison}, and the experimental comparison is detailed in section \ref{exp:experiment_comparison}.  

\begin{table}[h]
	\begin{center}
		\caption{Parameter of the Humanoid Robot}
		\begin{tabular}{c|c|c|c}
			\hline
			$m_1$ & Mass of shank &6.759 &Kg\\
			\hline
			$m_2$ & Mass of thigh &3.0426&Kg\\
			\hline
			$m_3$ & Mass of torso &15.04579&Kg\\
			\hline  
			$l^x_1$ & Length of shank &0.3538&m\\
			\hline
			$l^x_2$ & Length of thigh &0.367&m\\
			\hline
			$l^x_3$ & Length of torso &0.65&m\\
			\hline 
		\end{tabular}\label{table:parameter_of_humanoid_robot}
	\end{center}
\end{table}

\subsection{Design Example}\label{sec:gait_example}
\begin{figure}[!ht]
	\centering
	\includegraphics[scale=0.33]{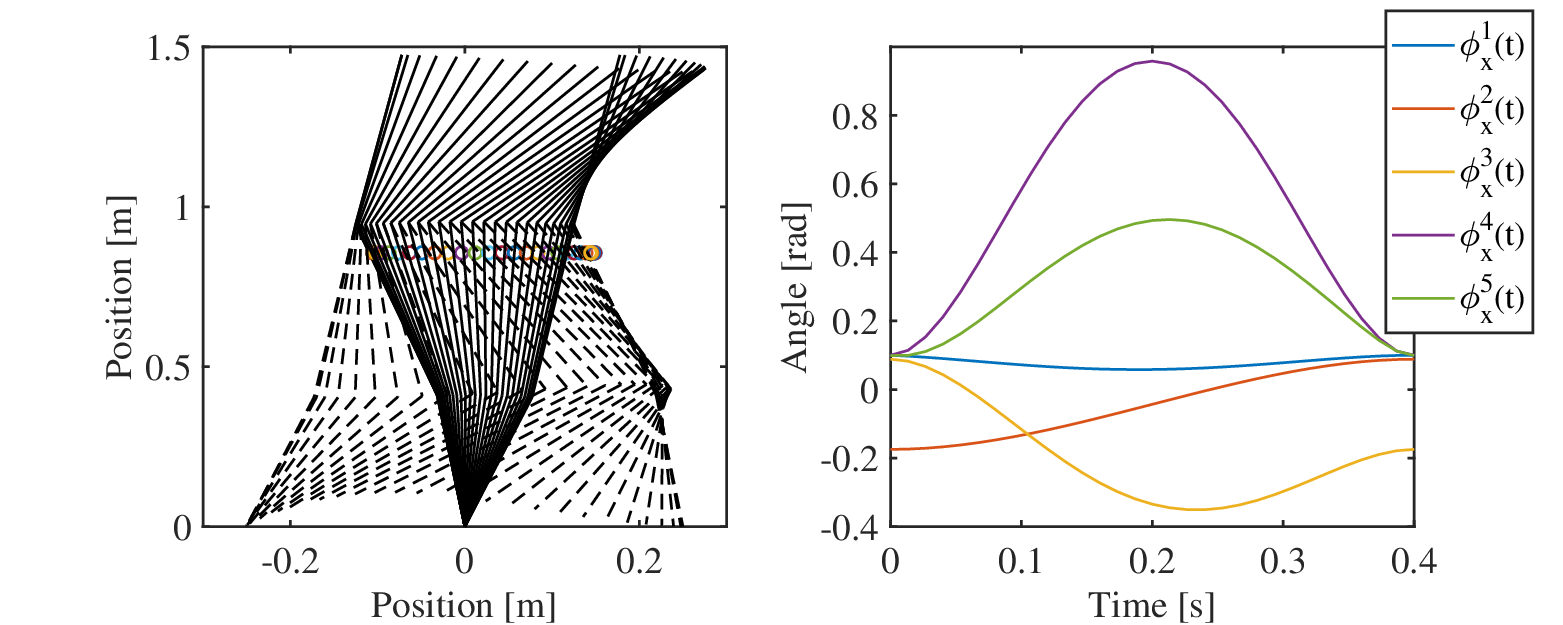}
	\caption{Stick animation of robot taking one step from left to right (left) and desired trajectories for X-model (right).}
	\label{fig_gait_design_x}
\end{figure}

\begin{figure}[!ht]
	\centering
	\includegraphics[scale=0.33]{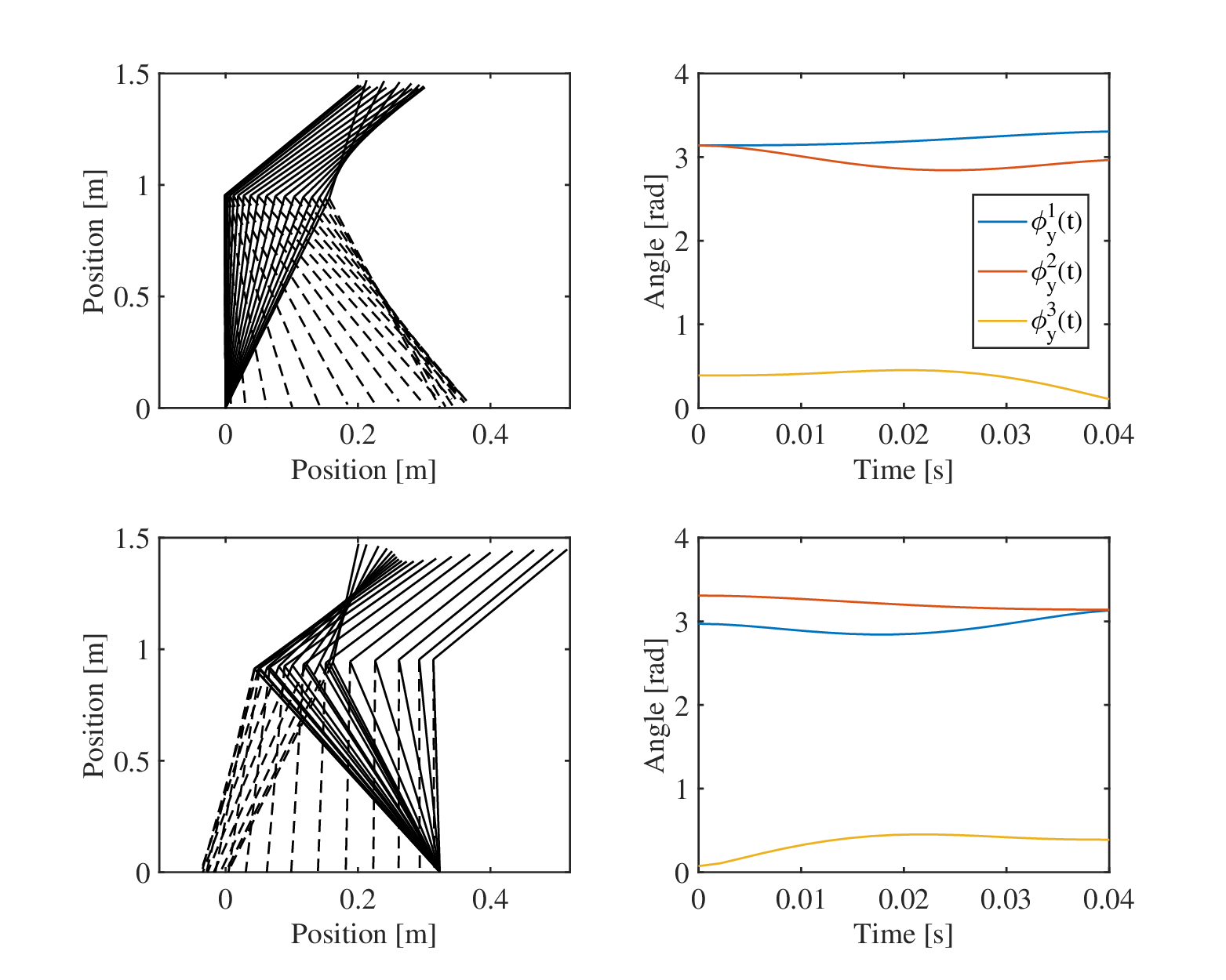}
	\caption{Stick animation of robot taking one step from left to right (left) and desired trajectories for Y-model (right).}
	\label{fig_gait_design_y}
\end{figure}

\begin{figure}[htbp]
	\centering
	\includegraphics[scale=0.25]{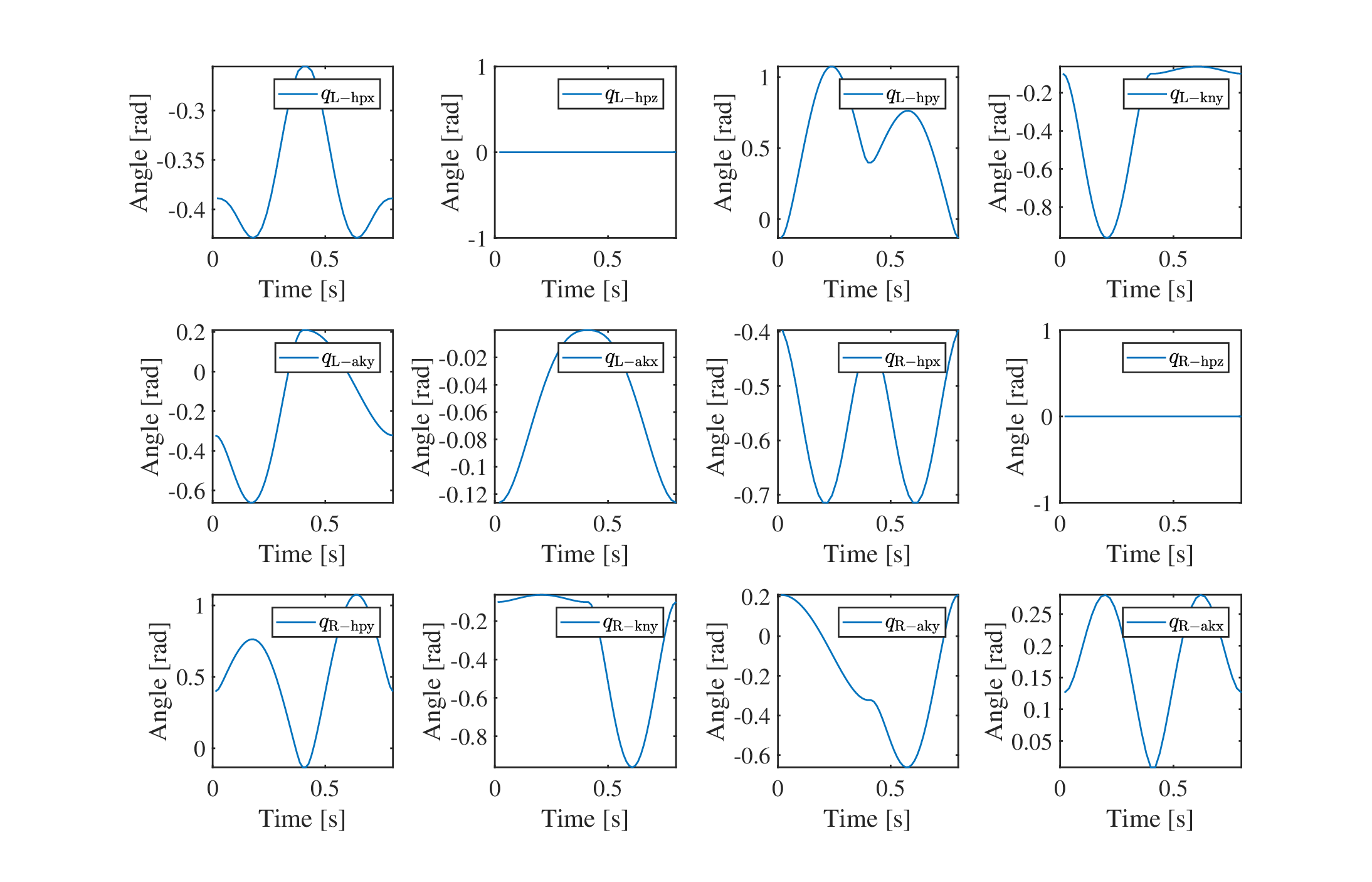}
	\caption{Joint trajectory of the humanoid robot generated from the joint trajectories of the $X$ model and the $Y$ model.}
	\label{fig_gait_robot_trajectory}
\end{figure}

First, we design the gait of $X$-model. We specify the step size $L = 0.25$ m and walking speed
$v = 0.625$ m/s. Choose $z_0 = 0.856$ m and the Bézier polynomials order $N = 4$. By applying the algorithm \ref{alg:search}. The stick animation and the desired trajectories is shown in Fig. \ref{fig_gait_design_x}. The parameter of Bézier polynomials for $\phi^x(t)$ is given in Table \ref{table:parameterX} of Appendix. As illustrated in Fig. \ref{fig_gait_design_x}, the step length of the robot is 0.25 m, and the angle of the torso in $X$-model undergoes periodic changes during the swing phase, with a swing period $T_{\rm{SSP}}$ = 0.4 s. To design the gait of $Y$-model, we specify the step size $L = 0.32$ m and walking speed $v = 0.8$ m/s. Choose $z_0 = 0.856$ m. By applying the algorithm \ref{alg:search1}. The stick animation and the desired trajectories is shown in Fig. \ref{fig_gait_design_y}. The parameter of Bézier polynomials for $\phi^y(t)$ is given in Table \ref{table:parameterY} of Appendix. Similar to Fig. \ref{fig_gait_design_x}, the step length of the $Y$-model is 0.32 m, and the angle of the torso in the $Y$-model undergoes periodic changes during the two swing periods, $2T_{\rm{SSP}}$. According to Table \ref{table:relationship_Decopule_model}, the robot's joint trajectory can be calculated from the joint trajectories of the $X$ model and the $Y$ model. Fig. \ref{fig_gait_robot_trajectory} shows the joint trajectory of the robot generated from the joint trajectories of the $X$ model and the $Y$ model within one gait cycle ($2T_{\rm{SSP}}$).

To verify the effectiveness of the gait design method, a velocity range of $-1.2$ m/s $\le V_x \le$ $1.2$ m/s and $-0.4$ m/s $\le V_y \le 0.4$ m/s is selected, with $K$ points uniformly sampled within this range. The swing period is set to $T_{\rm{SSP}} = 0.4$ s, and gait design was conducted for the $X$ model and the $Y$ model. The center of mass height $z_0$ is set to $0.825$ m. The total computation time for gait planning of the $X$ model and the $Y$ model is shown in Table \ref{table:consumption}. Table \ref{table:consumption} indicates that the proposed gait planning method can complete the robot's gait planning within 1 ms, meeting the requirements for real-time gait planning.


\begin{table}[h]
	\begin{center}
		\caption{The time consumption of the gait design}
		\begin{tabular}{c|c|c}
			\hline
			$K$ & Algorithm 1 & Algorithm 2\\
			\hline
			$10$ & 0.084587 & 0.006885\\
			\hline
			$100$ & 0.597201 & 0.010129\\
			\hline
			$1000$ & 5.823315 & 0.027321\\
			\hline
		\end{tabular}\label{table:consumption}
	\end{center}
\end{table}

\subsection{Simulation Comparison}\label{sec:performance_comparison}
\begin{figure}[!ht]
	\centering
	\includegraphics[scale=0.35]{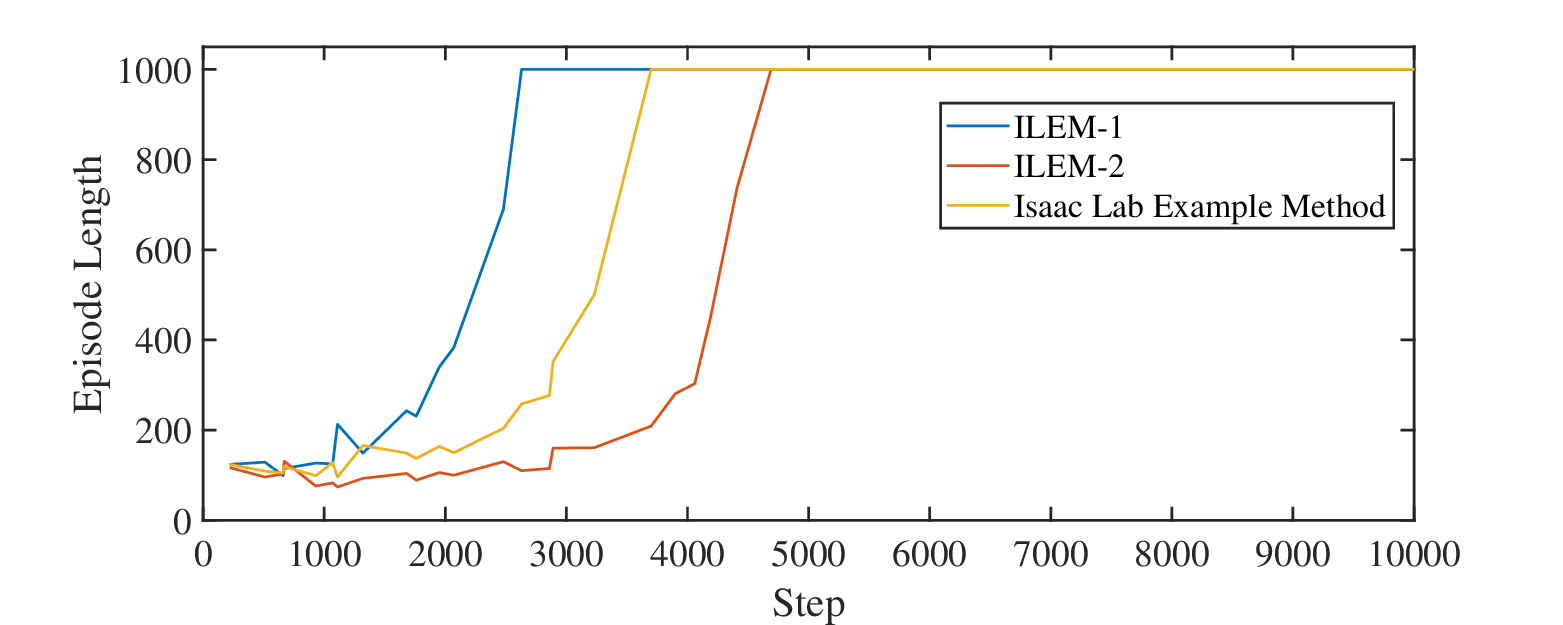}
	\caption{The average number of robot survivals at different training steps (total number: 1000).}
	\label{fig_robot_survivals}
\end{figure}

\begin{figure}[!ht]
	\centering
	\includegraphics[scale=0.30]{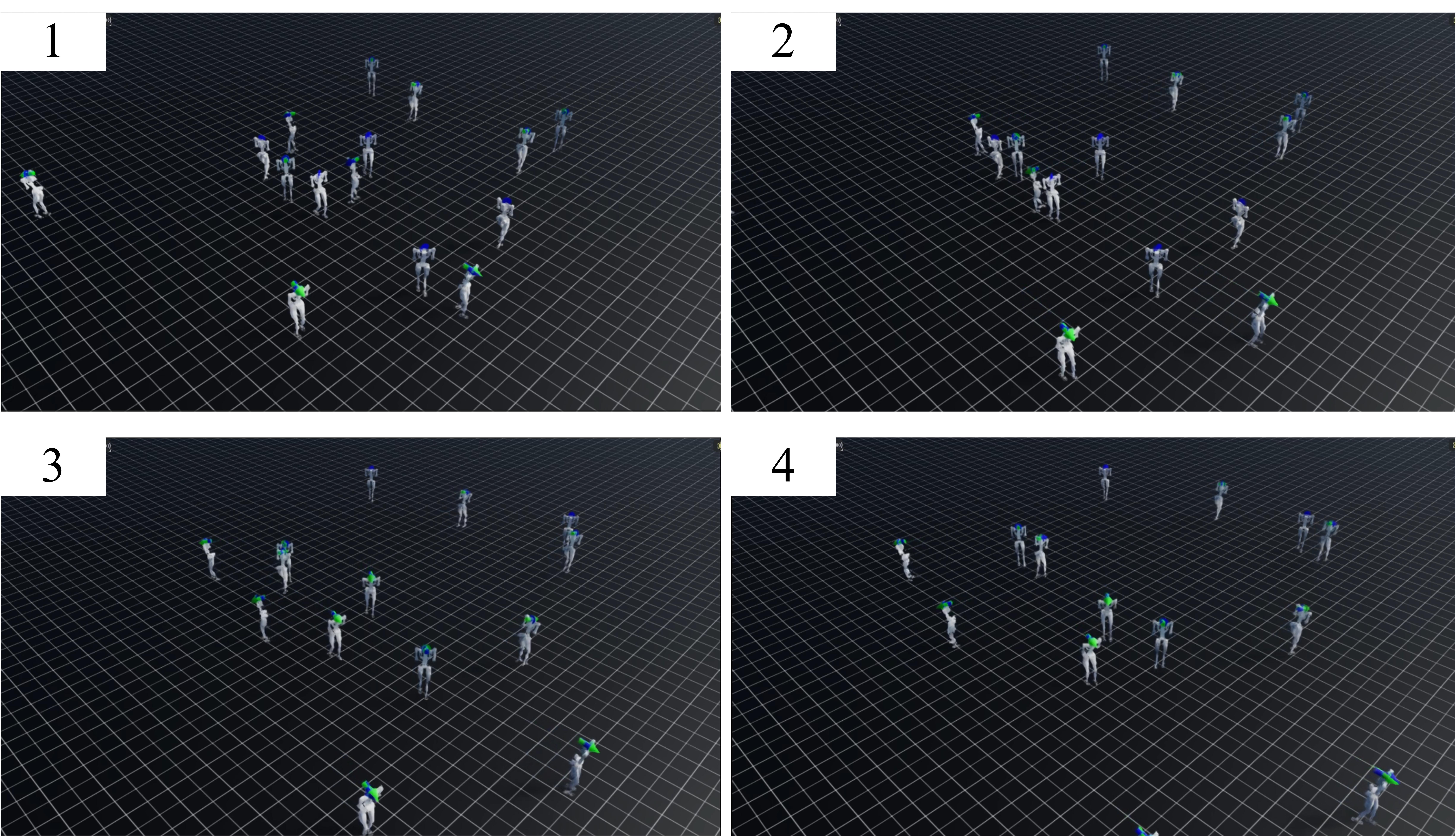}
	\caption{Screenshot of multiple robots' movement in the video after 200,000 steps of training using ILEM-1.}
	\label{fig_train_1}
\end{figure}

\begin{figure}[!ht]
	\centering
	\includegraphics[scale=0.30]{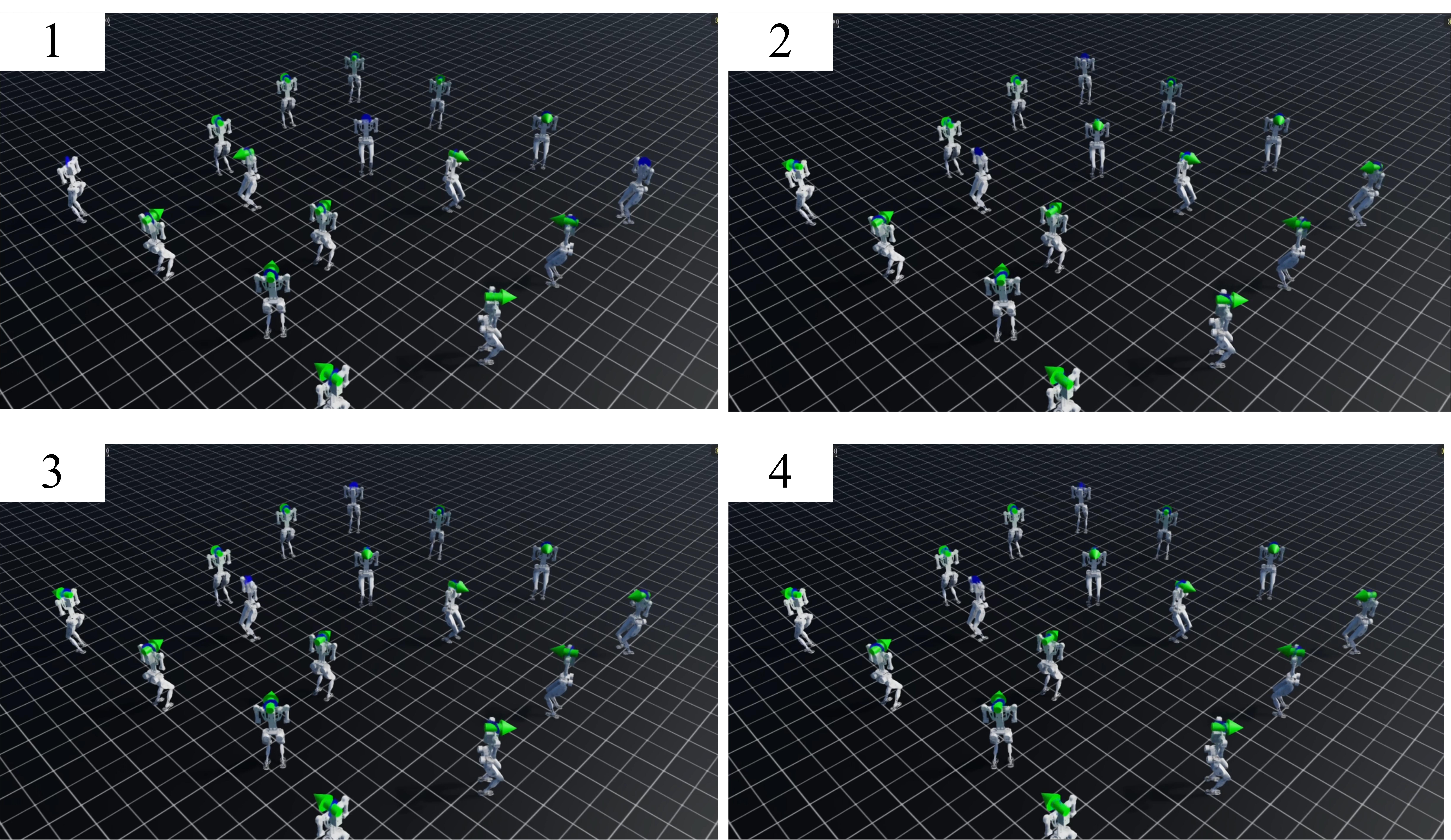}
	\caption{Screenshot of multiple robots' movement in the video after 200,000 steps of training using Isaac Lab Example Method.}
	\label{fig_train_2}
\end{figure}

To verify the effectiveness of the proposed reward composition, we train the robot in the Isaac Sim reinforcement learning environment for performance comparison. The training is deployed on a GPU equipped with the NVIDIA GeForce RTX 4090. The robot is trained using Isaac Lab example method (ILEM) \cite{mittal2023orbit} and the corresponding code can be found in the Isaac Lab repository. We compare the Isaac Lab example method with a method composed of ILEM and three reward compositions (ILEM-1), as well as a method composed of ILEM with the first two reward compositions (ILEM-2), to verify the effectiveness of the proposed reward composition. Fig. \ref{fig_robot_survivals} shows the evolution curve of the robot's survival duration during training, which directly reflects the balancing capability of the control strategy. As illustrated in Fig. \ref{fig_robot_survivals}, the robot trained with ILEM-1 achieves balance in a shorter time, and ILEM-1 is more effective than both ILEM-2 and the Isaac Lab Example Method. Fig. \ref{fig_train_1} and Fig. \ref{fig_train_2} show screenshots of multiple robots' movements in the video after 200,000 steps of training using ILEM-1 and the Isaac Lab Example Method, respectively. As illustrated in Fig. \ref{fig_train_2}, the robot using the Isaac Lab Example Method remains stationary, despite maintaining balance in the simulation environment, when the given velocity command is set to 0.625 m/s. Unlike the Isaac Lab Example Method, the robot using ILEM-1 is able to track the given velocity command with the specified periodic gait. This result demonstrates the effectiveness of the gait planner and the proposed reward composition within the reinforcement learning framework.

\begin{figure}[!ht]
	\centering
	\includegraphics[scale=0.35]{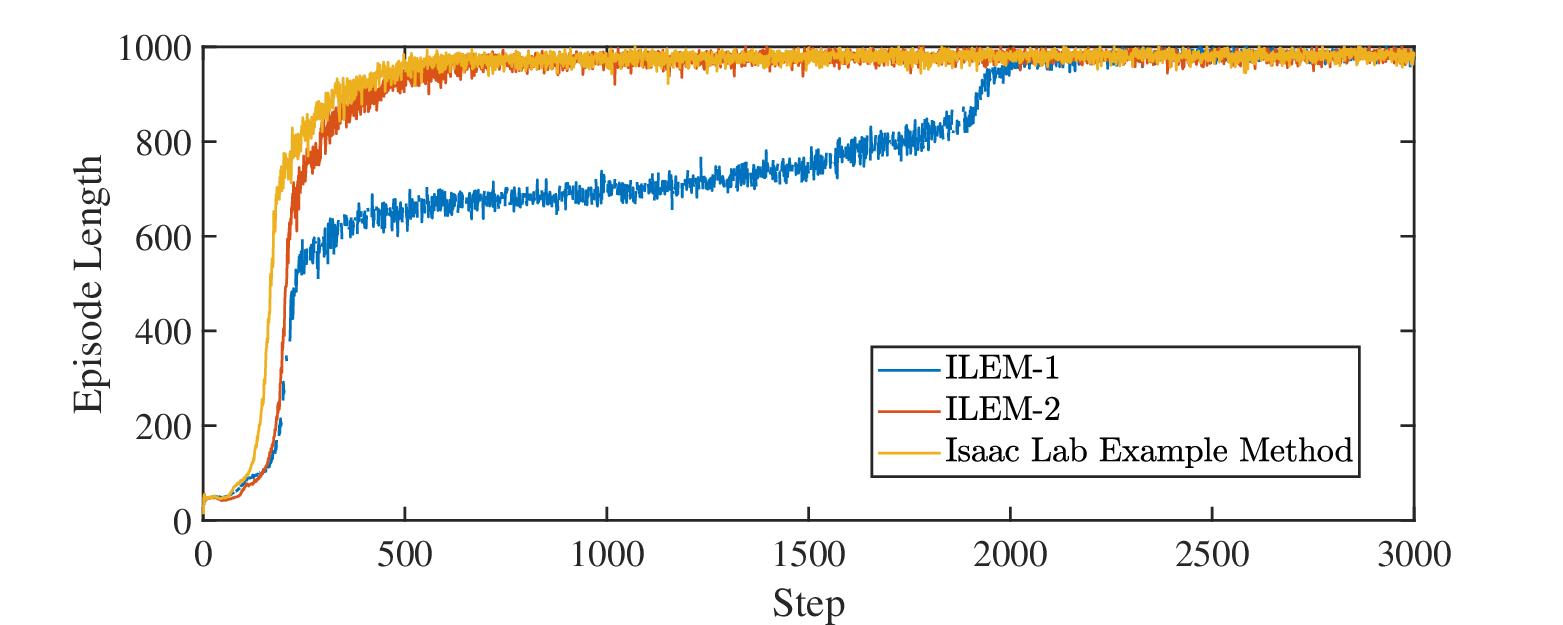}
	\caption{The average number of robot survivals at different training steps (total number: 1000).}
	\label{fig_robot_survivals_xy}
\end{figure}

\begin{figure}[!htbp]
	\centering
	\includegraphics[scale=0.48]{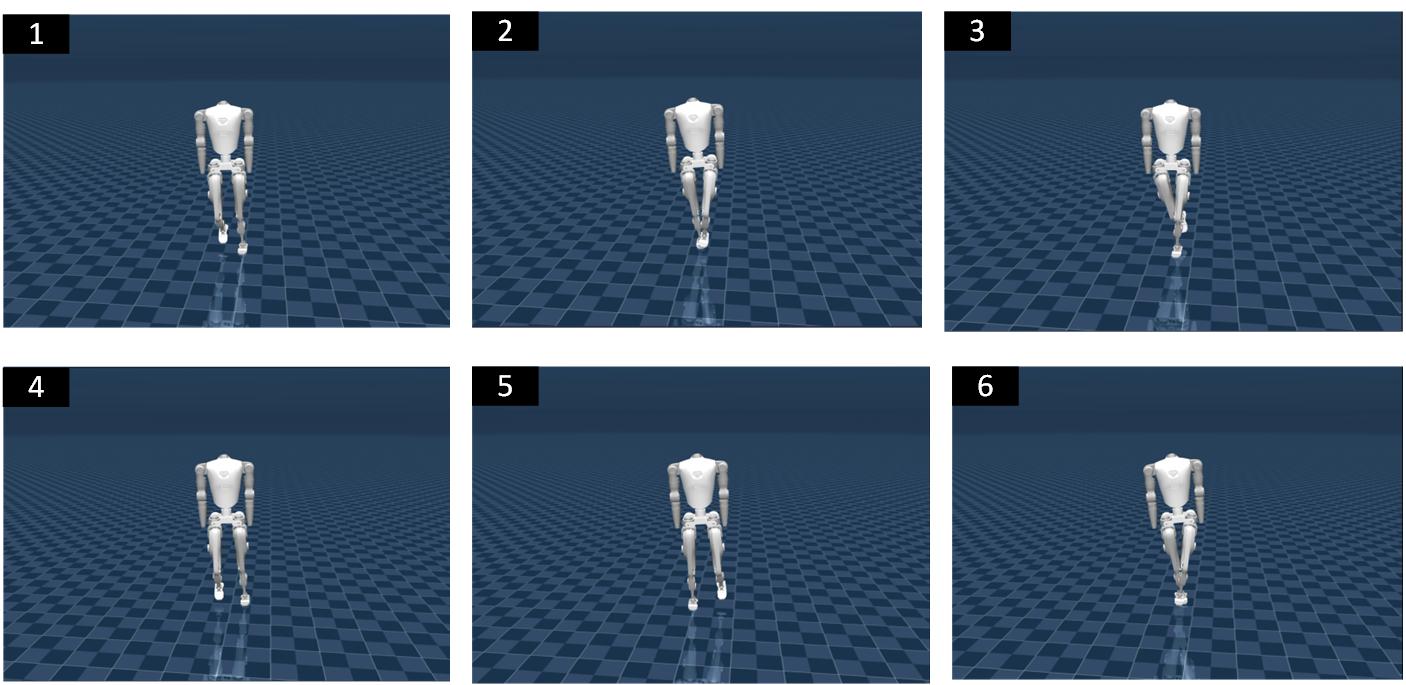}
	\caption{The results after training with ILEM-1 for 12,000 steps and deploying the network in the MuJoCo simulator.}
	\label{fig_train_biped}
\end{figure}

\begin{figure}[!ht]
	\centering
	\includegraphics[scale=0.36]{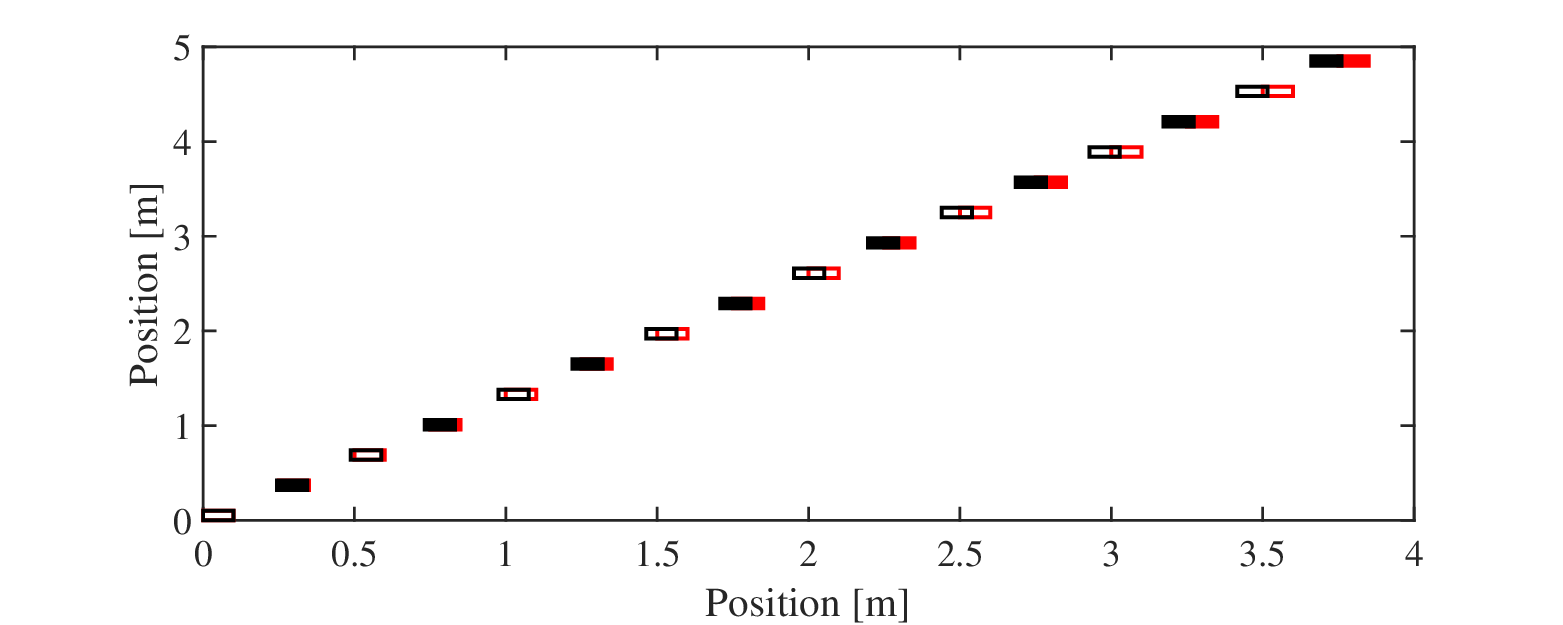}
	\caption{The footfall positions of the robot (solid rectangles: left leg footfall points, dashed rectangles: right leg footfall points; red rectangles: expected footfall points, black rectangles: actual footfall points).}
	\label{fig_gait}
\end{figure}

In order to further validate the effectiveness of the proposed framework, the velocities of the robot in the $x$ and $y$ axis directions are set to 0.625 m/s and 0.8 m/s, respectively. Fig. \ref{fig_robot_survivals_xy} presents the evolution curve of the robot's sustained standing time during training. As shown in Fig. \ref{fig_robot_survivals_xy}, the robot trained with ILEM-1 achieves balance in a shorter time, and its performance is significantly better than that of the original Isaac Lab example method. Fig. \ref{fig_train_biped} shows the simulation results after training with ILEM-1 for 12,000 steps, deployed in the MuJoCo simulator. Fig. \ref{fig_gait} corresponds to the expected and actual footfall locations in this simulation. Both Fig. \ref{fig_train_biped} and Fig. \ref{fig_gait} indicate that the robot is able to perfectly track the expected position along the $y$ axis, and shows good velocity tracking performance along the $x$ axis, demonstrating the effectiveness of the framework in utilizing the decoupled model for robot velocity planning.

\subsection{Experiment Comparison}\label{exp:experiment_comparison}
\begin{figure}[!ht]
	\centering
	\includegraphics[scale=0.72]{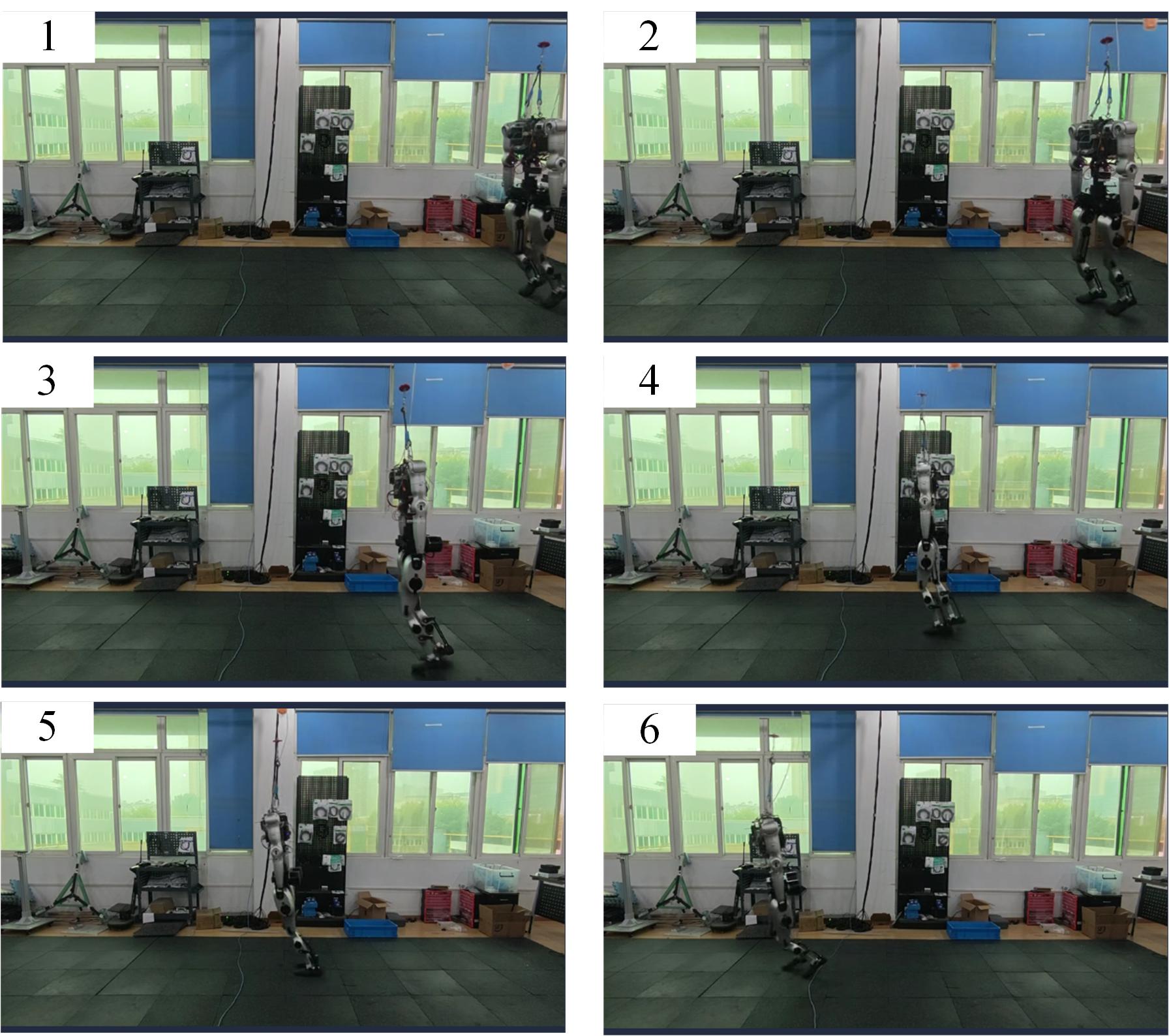}
	\caption{Screenshot of the real robot’s movement using the network trained with the Isaac Lab example method.}
	\label{fig_exp_1}
\end{figure}

\begin{figure}[!ht]
	\centering
	\includegraphics[scale=0.72]{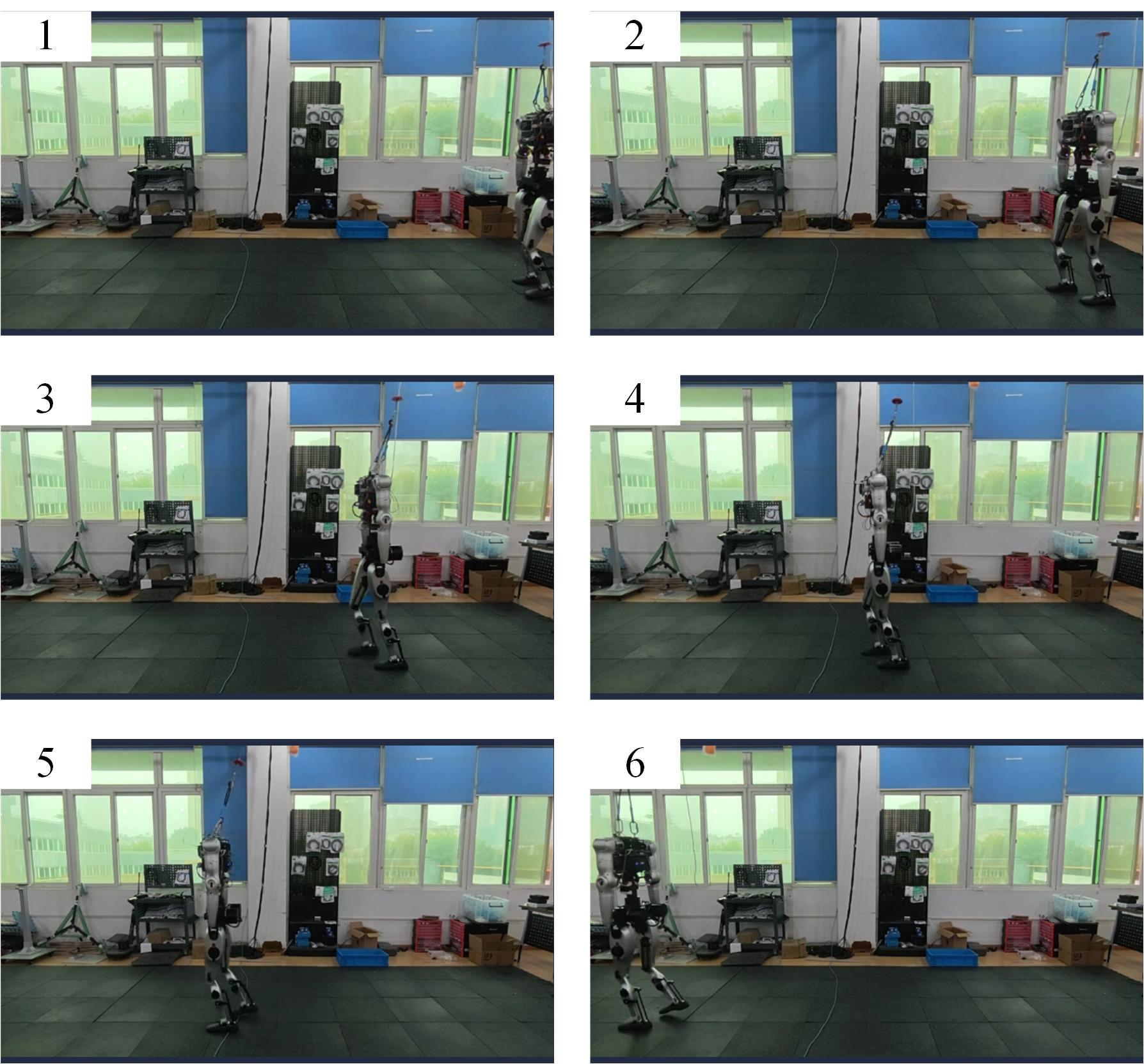}
	\caption{Screenshot of the real robot’s movement using the network trained with the ILEM-1.}
	\label{fig_exp_3}
\end{figure}

To evaluate the locomotion performance of the real robot, we conduct two experiments using networks trained with the ILEM-1 and Isaac Lab example methods. As illustrated in Fig. \ref{fig_exp_1}, the robot trained using the Isaac Lab example method was unable to walk at the specified speed, with its gait cycle detailed in Table \ref{table:gait_exp_1}.
Table \ref{table:gait_exp_1} shows that the robot utilizing the Isaac Lab example method does not achieve the specified gait cycle.
In contrast, the robot trained with the ILEM-1 method successfully achieved movement at the specified velocity, as demonstrated in Fig. \ref{fig_exp_3}, and its gait cycle is presented in Table \ref{table:gait_exp_3}. These results highlight the effectiveness of the proposed reward composition, confirming that the robot trained with the ILEM-1 method can reliably move according to the specified speed and the provided gait cycle.

\begin{small}
\begin{table}[htbp]
	\begin{center}
		\caption{Training the robot's gait cycle in the experiment using the Isaac Lab example method}
		\begin{tabular}{cccccccc}
			\hline
			$n$-th cycle & 1 & 2& 3& 4& 5& 6& 7\\
			\hline
			gait cycle(s)& 0.372 & 0.362&0.371 &0.287 &0.347 &0.119 &0.206 \\
			\hline
			\hline
			$n$-th cycle &8 & 9 & 10& 11&12& 13& 14 \\
			\hline
			gait cycle(s) &0.181 & 0.211 & 0.204&0.253 &0.208 &0.213 &0.171 \\
			\hline
		\end{tabular}\label{table:gait_exp_1}
	\end{center}
\end{table}
\end{small}

\begin{table}[!htbp]
	\begin{center}
		\caption{Training the robot's gait cycle in the experiment using the ILEM-1 method (Designed gait cycle: 0.8s)}
		\begin{tabular}{cccccccc}
			\hline
			$n$-th cycle & 1 & 2& 3& 4& 5& 6& 7\\
			\hline
			gait cycle(s) & 0.732 & 0.817&0.802 &0.782 &0.798 &0.81 &0.79 \\
			\hline
			\hline
			$n$-th cycle &8& 9 & 10& 11&12& 13& 14\\
			\hline
			gait cycle(s) &0.8& 0.82 & 0.78&0.79 &0.81 &0.82 &0.80 \\
			\hline
		\end{tabular}\label{table:gait_exp_3}
	\end{center}
\end{table}

\section{Conclusions}\label{sec:conclusion}
This paper introduces a novel approach for learning periodic bipedal gait in humanoid robots, which combines a dynamic gait planner with a reinforcement learning framework. By decoupling the 3D robot model into two 2D models and approximating them as H-LIP for trajectory planning, the gait planner effectively generates the desired joint trajectories. The incorporation of a reward composition within the reinforcement learning framework significantly reduces learning time while enhancing locomotion performance. The examples of gait design, simulation comparisons, and experimental evaluations demonstrate the effectiveness of the proposed method, providing a promising solution for improving bipedal gait in humanoid robots.

\appendix
Let $r^x_i$ for $i = 1,2,3,4$ represent the distances between the COM of the link and the corresponding joint for X-model. Then, $h^x_i$ and $v^x_i$ can be given by
\begin{align}
	h^x_1 =& r^x_1\sin(q^x_1 + q^x_2),\;\;
	v^x_1 = r^x_1\cos(q^x_1 + q^x_2),\nonumber\\
	h^x_2 =& l^x_1\sin(q^x_1 + q^x_2) + r^x_2\sin(q^x_2),\nonumber\\
	v^x_2 =& l^x_1\cos(q^x_1 + q^x_2) + r^x_2\cos(q^x_2),\nonumber\\
	h^x_3 =& l^x_1\sin(q^x_1 + q^x_2) + l^x_2\sin(q^x_2) -r^x_3\sin(q^x_3),\nonumber\\
	v^x_3 =& l^x_1\cos(q^x_1 + q^x_2) + l^x_2\cos(q^x_2) -r^x_3\cos(q^x_3),\nonumber\\
	h^x_4 =& l^x_1\sin(q^x_1 + q^x_2) + l^x_2\sin(q^x_2) -l^x_2 \sin(q^x_3) \nonumber\\&- r^x_4 \sin(q^x_3 + q^x_4),\nonumber\\
	v^x_4 =& l^x_1\cos(q^x_1 + q^x_2) + l^x_2\cos(q^x_2) -l^x_2 \cos(q^x_3) \nonumber\\ &- r^x_4 \cos(q^x_3 + q^x_4),\nonumber\\
	h^x_5 =& l^x_1\sin(q^x_1 + q^x_2) + l^x_2\sin(q^x_2) + r^x_5 \sin(q^x_5),\nonumber\\
	v^x_5 =& l^x_1\cos(q^x_1 + q^x_2) + l^x_2\cos(q^x_2) + r^x_5 \cos(q^x_5).\nonumber
\end{align}

Let $r^y_i$ for $i = 1,2,3$  represent the distances between the COM of the link and the corresponding joint for Y-model. Then, $h^y_i$ and $v^y_i$ can be given by
\begin{align}
	h^y_1 =& r^y_1 \sin(q^y_2 - \pi),\quad h^y_2 = l^y_1\sin(q^y_2 - \pi) + r^y_2\sin(\pi - q^y_1),\nonumber\\
	h^y_3 =& l^y_1\sin(q^y_2 - \pi) + r^y_3\sin(q^y_3),\nonumber\\
	v^y_1 =& r^y_1 \cos(q^y_2 - \pi),\quad v^y_2 = l^y_1\cos(q^y_2 - \pi) + r^y_2\cos(\pi - q^y_1),\nonumber\\
	v^y_3 =& l^y_1\cos(q^y_2 - \pi) + r^y_3\sin(q^y_3).\nonumber
\end{align}

The parameter of Bézier polynomials for $\phi^x(t)$ and $\phi^y(t)$ are given in Tables \ref{table:parameterX} and \ref{table:parameterY}, respectively.

\begin{table}[!h]
	\begin{center}
		\caption{The parameter of Bézier polynomials for $\phi^x(t)$}
		\begin{tabular}{c|c|c|c|c|c|c|c}
			\hline
			$\alpha^1_0$ & 0.1 &$\alpha^1_1$&0.084&$\alpha^1_2$&0 &$\alpha^1_3$&0.1\\
			\hline
			$\alpha^1_4$ & $0.1$ &$\alpha^2_0$&-0.175&$\alpha^2_1$&-0.175 &$\alpha^2_2$&-0.043\\
			\hline
			$\alpha^2_3$ & 0.088 &$\alpha^2_4$&0.088&$\alpha^3_0$&0.088 &$\alpha^3_1$&0.088\\
			\hline
			$\alpha^3_2$ &-0.82 &$\alpha^3_3$&-0.174&$\alpha^3_4$& -0.174 &$\alpha^4_0$ &0.1\\
			\hline
			$\alpha^4_1$ & 0.1 &$\alpha^4_2$& 2.4 &$\alpha^4_3$& 0.0835 &$\alpha^4_4$&0.1\\
			\hline
		\end{tabular}\label{table:parameterX}
	\end{center}
\end{table}

\begin{table}[!h]
	\begin{center}
		\caption{The parameter of Bézier polynomials for $\phi^y(t)$}
		\begin{tabular}{c|c|c|c|c|c|c|c}
			\hline
			$\beta^1_0$ & 3.14 &$\beta^1_1$&3.14&$\beta^1_2$&3.14 &$\beta^1_3$ &3.31\\
			\hline
			$\beta^1_4$ & 3.30 &$\beta^2_0$&3.14&$\beta^2_1$&3.14 &$\beta^2_2$&2.512\\
			\hline
			$\beta^2_3$ &2.97&$\beta^2_4$&2.97&$$& &\\
			\hline
		\end{tabular}\label{table:parameterY}
	\end{center}
\end{table}

\bibliographystyle{IEEEtran}
\bibliography{IEEEabrv,IEEEexample}
\end{document}